\documentclass[10pt,twocolumn,letterpaper]{article}

\usepackage[pagenumbers]{cvpr} 

\usepackage{soul}

\definecolor{cvprblue}{rgb}{0.21,0.49,0.74}
\usepackage[pagebackref,breaklinks,colorlinks,allcolors=cvprblue]{hyperref}

\usepackage{hyperref}
\usepackage{url}
\usepackage{xcolor}
\usepackage{booktabs}
\usepackage{pifont}
\usepackage{makecell}
\usepackage{placeins}
\usepackage{graphicx}
\usepackage{booktabs}
\usepackage{tcolorbox}
\usepackage{stfloats}

\usepackage{colortbl}
\usepackage{xcolor}
\usepackage{makecell}
\usepackage{multirow}
\usepackage{float}

\usepackage{url}

\definecolor{bestgreen}{HTML}{4BA36A}   
\definecolor{secondgreen}{HTML}{B7DBA8} 

\def\paperID{3857} 
\def\confName{CVPR}
\def\confYear{2026}

\title{ConsID-Gen:\\View-Consistent and Identity-Preserving Image-to-Video Generation}

\author{
    Mingyang Wu$^{1}$, 
    Ashirbad Mishra$^{2}$, 
    Soumik Dey$^{2}$,
    Shuo Xing$^{1}$, 
    Naveen Ravipati$^{2}$,\\
    Hansi Wu$^{2}$,
    Binbin Li$^{2}$,
    Zhengzhong Tu$^{1}$$^{\dagger}$\\[0.2cm]
    $^1$Texas A\&M University \quad
    $^2$eBay Inc.\\[0.2cm]
}

\begin{document}
\twocolumn[{
\renewcommand\twocolumn[1][]{#1}
\maketitle
\begin{center}
    \centering
    \captionsetup{type=figure}
    \includegraphics[width=1\textwidth]{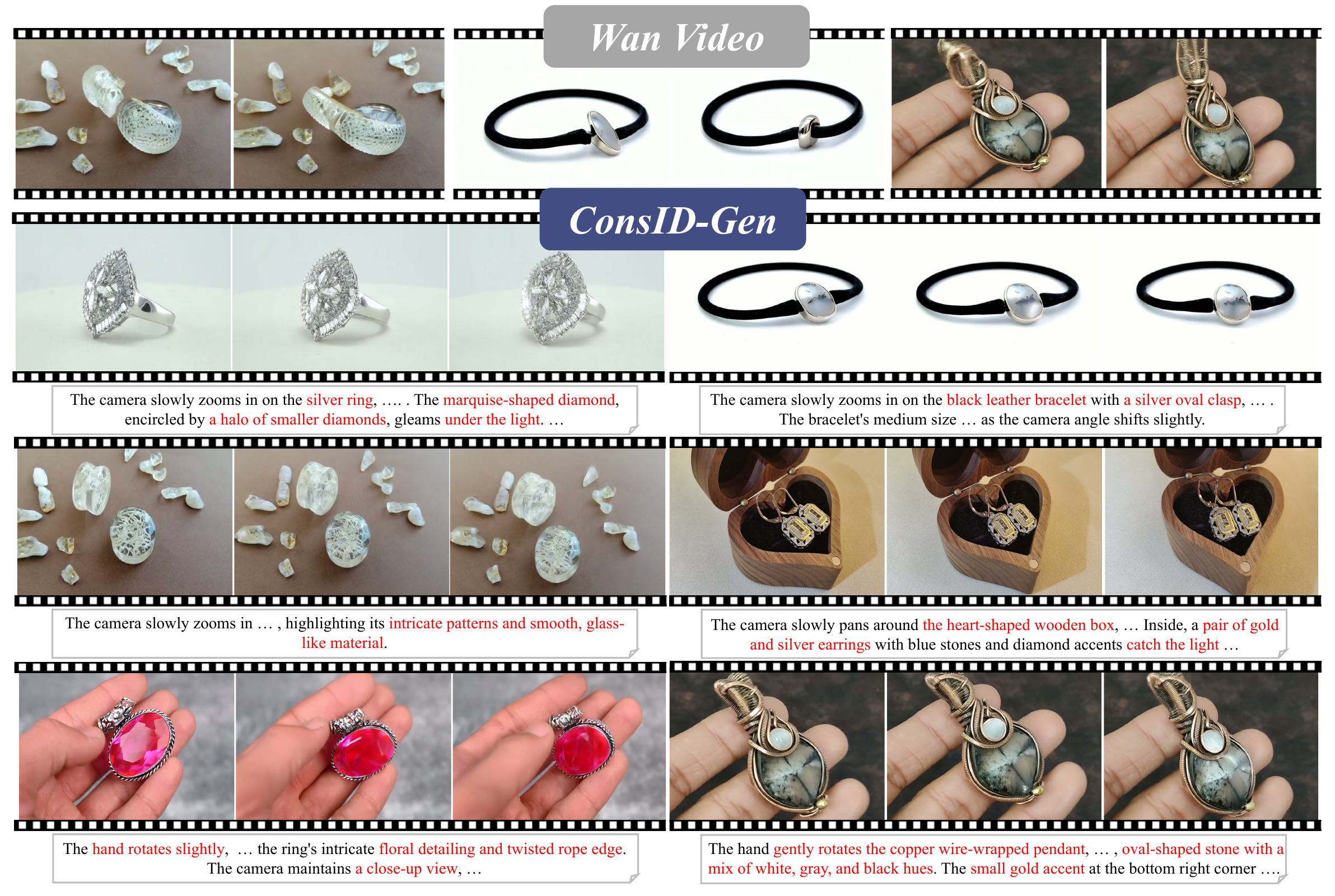}
    \vspace{-5mm}
    \captionof{figure}
    {
    \textbf{Examples Synthesized by ConsID-Gen.}
    Given a textual instruction and reference image containing rigid objects (\textit{i.e., rings, diamonds}), \textbf{ConsID-Gen} synthesizes realistic videos that faithfully preserve \emph{object identity} and maintain \emph{geometric consistency}. The initial row was generated by Wan~\cite{wan2025wan} using the same prompt. Attributes highlighted in \textcolor{red}{red} denote object properties specified in the instruction.
    }
\label{fig: teaser}
\end{center}
}]
\maketitle
\begingroup
  \renewcommand\thefootnote{}\footnote{$^\dagger$ Corresponding author: tzz@tamu.edu}
  \addtocounter{footnote}{-1}
\endgroup

\begin{abstract}
Image-to-Video generation (I2V) animates a static image into a temporally coherent video sequence following textual instructions, yet preserving fine-grained object identity under changing viewpoints remains a persistent challenge.
Unlike text-to-video models, existing I2V pipelines often suffer from appearance drift and geometric distortion, artifacts we attribute to the sparsity of single-view 2D observations and weak cross-modal alignment.
Here we address this problem from both data and model perspectives. 
First, we curate \textbf{ConsIDVid}, a large-scale object-centric dataset built with a scalable pipeline for high-quality, temporally aligned videos, and establish \textbf{ConsIDVid-Bench}, where we present a novel benchmarking and evaluation framework for multi-view consistency using metrics sensitive to subtle geometric and appearance deviations. 
We further propose \textbf{ConsID-Gen}, a view-assisted I2V generation framework that augments the first frame with unposed auxiliary views and fuses semantic and structural cues via a dual-stream visual–geometric encoder as well as a text–visual connector, yielding unified conditioning for a Diffusion Transformer backbone. 
Experiments across ConsIDVid-Bench demonstrate that ConsID-Gen consistently outperforms in multiple metrics, with the best overall performance surpassing leading video generation models like Wan2.1 and HunyuanVideo, delivering superior identity fidelity and temporal coherence under challenging real-world scenarios. 
We will release our model and dataset at \href{https://myangwu.github.io/ConsID-Gen}{https://myangwu.github.io/ConsID-Gen}.

\end{abstract}    
\section{Introduction}
\label{sec:intro}

Modern video generation models based on diffusion transformers (DiT)~\cite{hong2022cogvideo, kong2024hunyuanvideo, wan2025wan, zhou2024allegro, polyak2024moviegen} can synthesize high-resolution, temporally coherent videos from text prompts, images, or both.
This progress is beginning to reshape applications in advertising~\cite{gao2025wans2vaudiodrivencinematicvideo}, entertainment~\cite{lumaai}, and digital content creation~\cite{cheng2025wan, kling, team2025kling}, where short, high-quality videos can now be synthesized rather than filmed~\cite{bai2025recammaster}.
Within this space, Image-to-Video (\textit{I2V}) generation~\cite{kong2024hunyuanvideo, wan2025wan, huang2025step} is especially appealing: given a single reference image and a textual instruction, an \textit{I2V} model animates a still frame into a temporally consistent, semantically rich video clip.
This capability is particularly valuable for product-centric scenarios where a single catalog photo must be turned into multiple compelling videos or hand-held showcases while preserving the exact appearance~\cite{peng2025open-sora2, heygen, Invideo}.

Despite this promise, preserving fine-grained object identity under changing viewpoints remains challenging.
Existing \textit{I2V} systems~\cite{ren2024consisti2v, kong2024hunyuanvideo, wan2025wan} frequently exhibit \textbf{appearance drift} or \textbf{geometric distortion}: identity shifts, object shape warps, parts merge or disappear, and materials or textures subtly change across frames.
As illustrated in Fig.~\ref{fig: teaser}, the glass gradually loses rigidity and appear to merge, violating the preservation of object-centric appearance.
This failure to maintain instance-level consistency is a major roadblock for deploying \textit{I2V} generation in real-world, high-stakes applications such as e-commerce, product advertising, and training videos.

\begin{figure}[t]
  \centering
  \includegraphics[width=\linewidth]{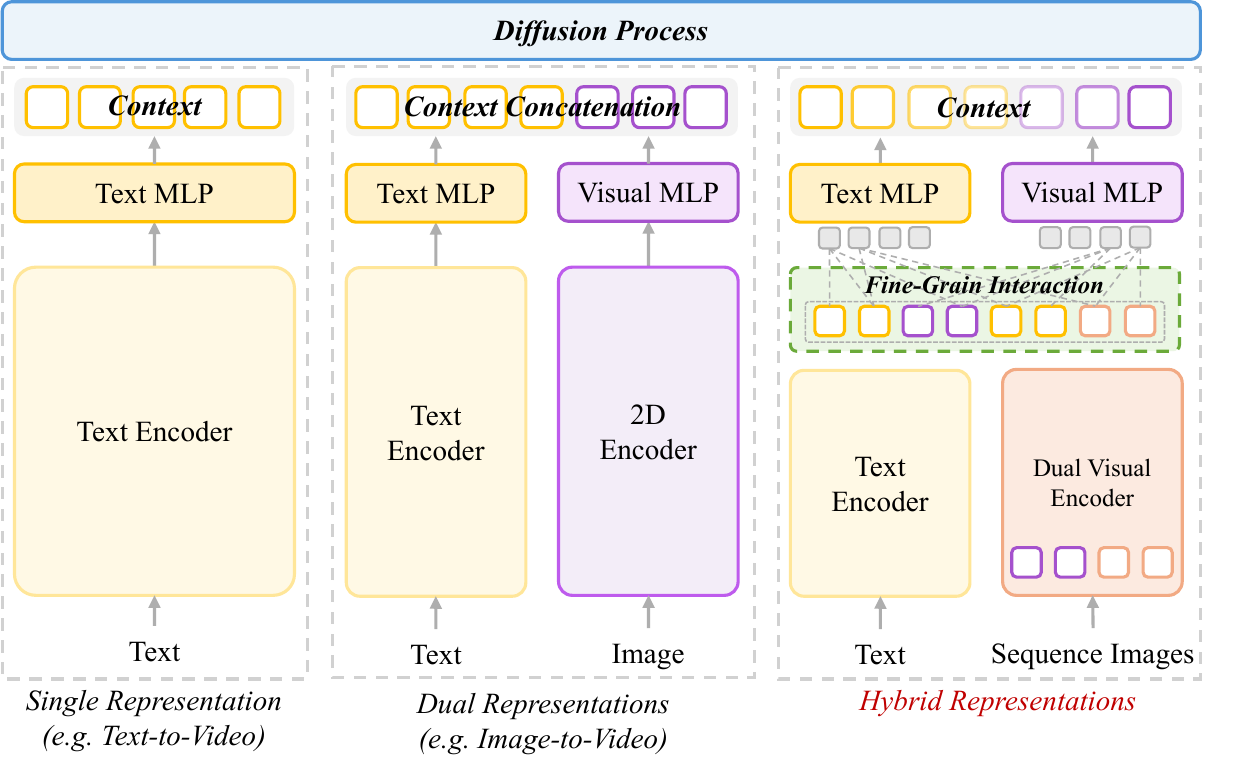}
  \vspace{-5mm}
  \caption{\textbf{Comparing Different Video Generation Paradigms.} Single-stream (\textit{T2V}) uses only text tokens as context. Dual-stream (\textit{I2V}) concatenates text and 2D visual tokens with limited interaction. Hybrid representations (\textit{Ours}) pre-align text and visual tokens via fine-grained interaction before projection.
  }
  \label{fig:compar_arch}
  \vspace{-4mm}
\end{figure}

Prior works~\cite{jeong2025track4gen} have explored explicit spatial supervision as a way to improve appearance drift; however, these methods are typically trained on small-scale curated datasets and evaluated on benchmarks~\cite{huang2024vbench} that emphasize semantic video quality rather than object identity.
Consequently, they provide limited insight into preserving consistent object geometry and appearance, and they often fail to generalize to real-world product scenarios.
More broadly, today's \textit{I2V} ecosystem suffers from two systemic limitations:
\textbf{1)} The available data is insufficient, where existing datasets rarely contain close-up, object-centric, multi-view videos that focus on identity continuity across space-time;
\textbf{2)} Model architectures are not structurally equipped to preserve identity. 
For instance, we find that \textit{T2V} models consistently outperform \textit{I2V} models in identity-preserving generation (e.g., CogVideoX-1.5~\cite{yang2024cogvideox}: \(95.77\%\!\rightarrow\!91.30\%\); Wan2.1~\cite{wan2025wan}: \(96.72\%\!\rightarrow\!91.84\%\); as shown in Table~\ref{tab:t2v_i2v_comparison}). We attribute this gap to a fundamental architectural issue where prevailing pipelines encode text and image inputs separately and only fuse them lately in the network (Fig.~\ref{fig:compar_arch}).

Motivated by these observations, here we systematically address the identity preservation issues in \textit{I2V} generation from both the data and model perspectives.
\ul{On the data side}, we build \textbf{ConsIDVid}, a large-scale object-centric dataset constructed through a scalable pipeline (Sec.~\ref{ssec:data-pipeline}) that selects high-quality, temporally aligned videos of rigid objects, and establish \textbf{ConsIDVid-Bench}, a dedicated benchmark that reframes \textit{I2V} evaluation as a multi-view consistency problem. 
Instead of relying on scene-level or frame-level scores, ConsIDVid-Bench incorporates geometry- and appearance-aware metrics explicitly designed to capture subtle distortions, shape inconsistencies, and within-object drift across viewpoints and time.

\ul{On the modeling side}, we propose \textbf{ConsID-Gen}, a view-assisted \textit{I2V} generation framework designed to explicitly encode appearance consistency and geometric stability. 
ConsID-Gen augments the single reference frame with unposed auxiliary views of the same object, allowing the model to recover richer structural cues to build a stable representation of object identity.
These visual inputs are processed through a dual-stream visual–geometric encoder that captures both semantic appearance features and multi-view geometry. 
A multimodal text–visual connector then aligns these cues with textual motion instructions to produce unified conditioning for a diffusion-based video backbone.
Our experimental results demonstrate that ConsID-Gen improves identity-preserving video generation. It delivers state-of-the-art identity fidelity on the proprietary subset and strong geometric preservation, achieving the lowest MEt3R (+30.2\%) on the proprietary set and the lowest Chamfer Distance (+7.26\%) on the public set.

\textbf{Our main contributions} are threefold: 
(i) A holistic I2V benchmark for identity preservation, with a diverse dataset and a novel multi-view evaluation suite;
(ii) ConsID-Gen introduces unified representation before diffusion, with multi-view guidance and improved cross-modal alignment;
(iii) Showing that ConsID-Gen outperforms open-source SOTAs in identity consistency and in human evaluation.
\section{Related Works}
\label{sec:related_works}

\subsection{Video Generation Models.}

\noindent\textbf{Text-Guided Video Generation.}
The generation of videos from textual descriptions has garnered significant scholarly interest in the past year, spurred by advancements ranging from Sora~\cite{openaisora} to MovieGen~\cite{polyak2024moviegen}, Gen-4~\cite{runway}, Sora2~\cite{openaisora2}, Kling~\cite{kling, team2025kling}, Veo 3~\cite{veo3}, and others. 
In particular, Sora, which synthesizes a temporal Variational Autoencoder (VAE) with a DiT backbone, represents a critical achievement that has stimulated extensive architectural research within open-source communities. 
Prominent studies such as CogVideo~\cite{hong2022cogvideo} and CogVideoX~\cite{yang2024cogvideox} utilize a three-dimensional variational autoencoder coupled with an expert Transformer, allowing for the generation of high-fidelity video content. HunyuanVideo~\cite{kong2024hunyuanvideo} and Mochi~1~\cite{mochi1} implement asymmetric architectures and comprehensive attention mechanisms to improve the alignment between textual and video data. 
Wan2.1~\cite{wan2025wan} enhances the model capacity, while Wan2.2~\cite{wan2025wan2.2} incorporates a sparse Mixture-of-Experts (MoE) approach, which delegates the diffusion process to specialized experts, thereby effectively capturing intricate motion dynamics.

\noindent\textbf{Text-Image-Guided Video Generation.}
Despite recent advances, text-only prompting in \textit{T2V} affords limited control over content and appearance.
A promising alternative is to extend pretrained video generators by modifying their architecture to incorporate image conditions. 
Within this paradigm, DynamiCrafter~\cite{xing2024dynamicrafter} and Moonshot~\cite{zhang2024moonshot} inject image embeddings via cross-attention layers. 
ConsistI2V~\cite{ren2024consisti2v} applies spatial–temporal attention to the first frame coupled with a frequency-aware noise initialization strategy to enhance temporal coherence.
SVD~\cite{blattmann2023stable} and CogVideoX~\cite{yang2024cogvideox} extend \textit{T2V} to \textit{I2V} by channel-wise concatenation of conditional latents with noise. 
Wan2.1~\cite{wan2025wan} adopts mask-guided conditioning and injects image embeddings via decoupled cross-attention.
Furthermore, such conditioning techniques are adapted for subject-to-video generation~\cite{liu2025phantom, fei2025skyreels, li2026skyreelsv3} and video editing~\cite{jiang2025vace} to ensure identity preservation and precise modification.

\subsection{Video Generation Evaluations.}

\noindent\textbf{Evaluations Metrics for Video Generation.}
With advances in generation, systematic evaluation of video quality has become increasingly crucial. Early works relied on distribution-based metrics such as Fréchet Video Distance (FVD)~\cite{unterthiner2018towards} and its variants~\cite{luo2024beyond}, which, despite widespread use, offer limited correspondence to human perception. 
Several \textit{T2V} evaluation benchmarks like VBench~\cite{huang2024vbench} provide structured, multi-dimensional evaluations focusing on fundamental visual attributes and prompt adherence, but their dependence on generic similarity models restricts fine-grained assessment. 
More recently, VLM-driven evaluators~\cite{he2024videoscore, zheng2025vbench2, bansal2025videophy} leverage inherent vision–language understanding to score intrinsic faithfulness; UVE~\cite{liu2025uve} further unifies this paradigm by prompting a VLM to perform both single-video rating and pairwise comparison under aspect-specific guidelines. 
Nevertheless, VLM-based approaches remain sensitive to prompt design and model bias.

\section{ConsIDVid Dataset \& Benchmark Curation}
\label{sec:dataset_benchmark}

\begin{figure*}[!t]
  \centering
  \includegraphics[width=\textwidth]{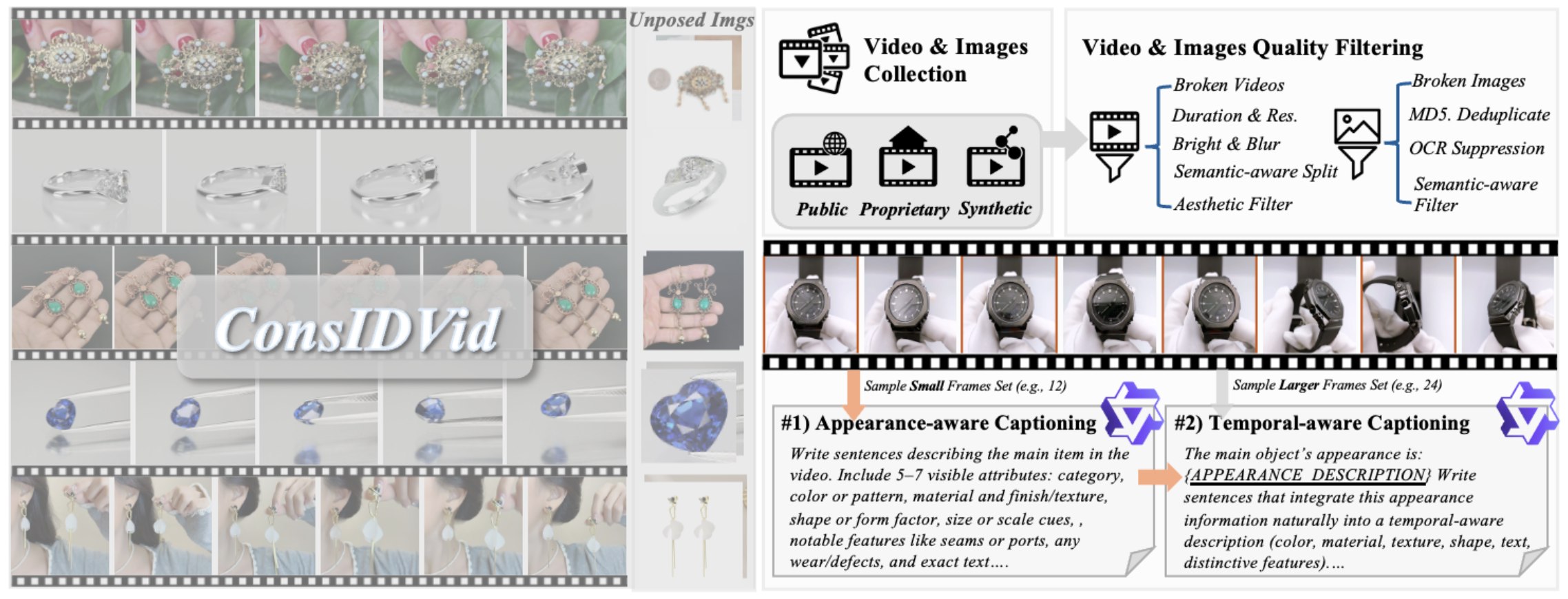}
  \vspace{-6mm}
  \caption{\textbf{Data Curation Pipeline.} We curate and synthesize videos from diverse sources, followed by an automated data curation pipeline to ensure visual and temporal quality. Video captions are produced by Qwen2.5-VL via a hierarchical captioning strategy.}
  \label{fig:data-curation-pipeline}
  \vspace{-4mm}
\end{figure*}

\begin{figure}[b]
  \centering
  \includegraphics[width=\columnwidth]{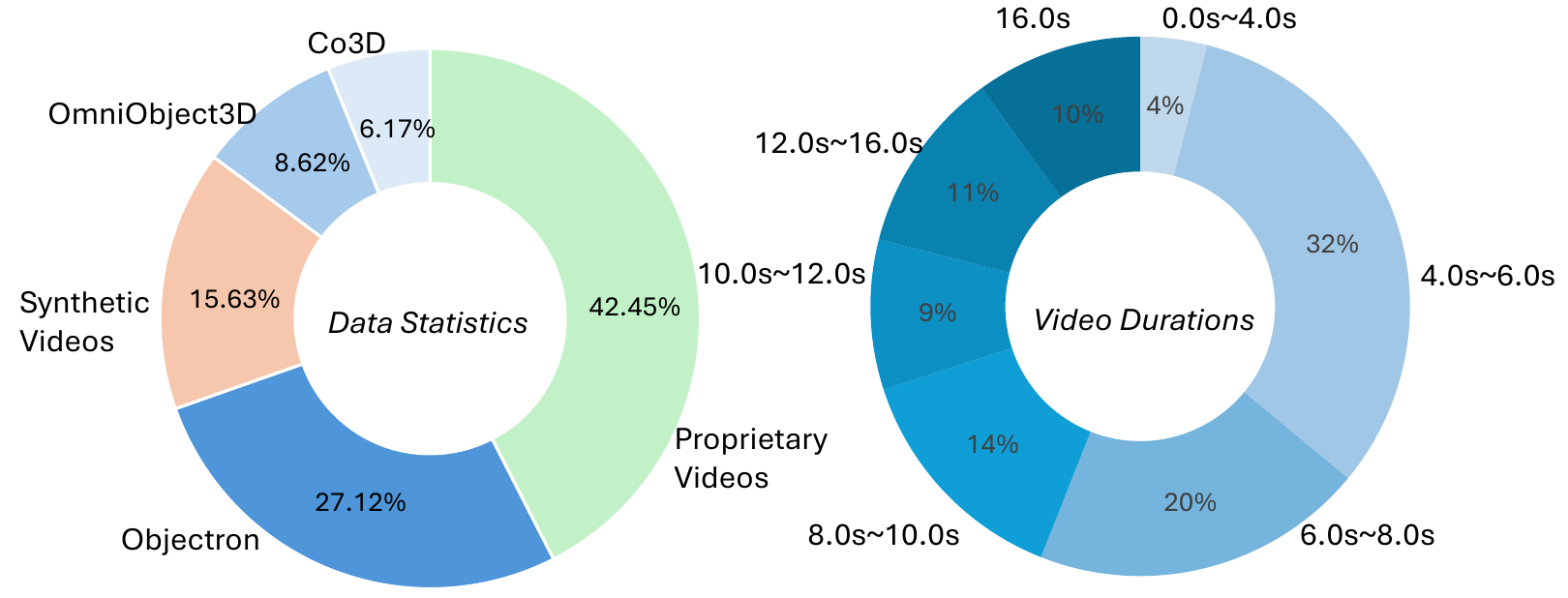}
  \vspace{-5mm}
  \caption{Statistics of video clips in ConsIDVid. The dataset includes diverse distributions of data source and video duration.}
  \label{fig:data-statistics}
\end{figure}

Prior methods~\cite{jeong2025track4gen} were trained on small, minimally curated appearance‑preserving datasets ($\sim$ 600 videos), which limits identity‑consistent modeling. In response, we present ConsIDVid, a large‑scale object‑centric, identity‑preserving video dataset curated via a scalable pipeline, together with an object‑preserving benchmark for standardized evaluation of \textit{I2V} models. We illustrate the data curation pipeline in Fig.~\ref{fig:data-curation-pipeline} and present its stats in Fig.~\ref{fig:data-statistics}.

\subsection{Video Collection}
\label{ss:video_collect}

To mitigate data scarcity, we curated a candidate dataset from three sources:
(i) existing object-centric datasets (\textit{Co3D}~\cite{reizenstein21co3d}, \textit{OmniObject3D}~\cite{wu2023omniobject3d}, \textit{Objectron}~\cite{ahmadyan2021objectron});
(ii) proprietary monocular videos; and
(iii) synthetic videos.
\textit{Co3D} provides in-the-wild, object-centric videos across 50 MS-COCO categories;
\textit{OmniObject3D} comprises 6{,}000 objects spanning 190 categories with accompanying real-world videos;
\textit{Objectron} offers $\sim$15{,}000 short clips across nine categories collected in 10~countries.

To further investigate the aspect of realism and practicality, 
we experiment on over 80~hours of object-centric monocular UGC from public e-commerce platforms,
where each clip primarily showcases a single product; many entries include unposed, multi-view images of the same item for instance-level supervision. 
We also synthesize object-centric sequences using a video generator conditioned on first and last keyframes, yielding temporally coherent clips suitable for identity-preserving training.

\subsection{Data Curation Pipeline}
\label{ssec:data-pipeline}

\noindent\textbf{Video Preprocessing.}
In the initial stage, we convert image sequences into standardized video clips and perform validity checks using FFmpeg. 
These steps remove a substantial fraction of unusable media at the outset of the pipeline.

\noindent\textbf{Video Quality Filtering.}
We implement a multi-faceted video quality filter to remove unsuitable content: 
\textbf{(i)} \textit{Duration and resolution}: each clip contains at least 81 frames and meets a minimum resolution of 320p;
\textbf{(ii)} \textit{Brightness and blur}: we prune the bottom/top 5\% of the luminance and Laplacian-variance distributions to remove under-/over-exposed and excessively blurred clips;
\textbf{(iii)} \textit{Semantics-aware splitting}: a two-stage procedure first detects shot boundaries and then stitches adjacent segments using frame-embedding similarity to correct over-segmentation, handle fade-in/out transitions and long uncut sequences, and reduce redundancy (cf.\ Panda-70M~\cite{chen2024panda});
\textbf{(iv)} \textit{Aesthetics}: we apply the LAION-5B~\cite{schuhmann2022laion} aesthetics predictor on 10 uniformly sampled frames to discard low-quality videos whose mean score is below 3.0.
To ensure scalability, proprietary videos are clustered and processed in batches under the same pipeline.

\noindent\textbf{Image Filtering.} 
We curate proprietary unposed multi-view object images with cascade image filters: \textbf{(i)} \textit{validity \& exact deduplication}: eliminate corrupt files and MD5 duplicates; \textbf{(ii)} \textit{OCR suppression}: discard images containing more than 30 detected characters; \textbf{(iii)} \textit{semantics-aware outlier removal}: apply CLIP-based~\cite{radford2021learning} reference matching to a curated outlier gallery and per-item embeddings clustered by DBSCAN, retaining the dominant cluster.

\subsection{Hierarchical Video Captioning}
\label{hierarchical_video_captioning}

The accuracy of captions is crucial for training video generation models. 
While Mixture-of-Multimodal-Experts captioning improves detail, its multi-model, multi-step inference is costly. By leveraging our object-centric dataset, we propose a two-stage hierarchical captioning protocol that produces fine-grained, temporally grounded video–text pairs with low computational overhead. We use Qwen2.5-VL~\cite{bai2025qwen2} as the captioner and uniformly sample frames.

\noindent\textbf{Stage~1: Appearance-aware Captioning.} 
From a small frame subset (e.g., 12), produce a caption restricted to the primary object’s visible attributes.
The prompt restricts content to 5–7 concrete cues: category; color/pattern; material/finish/texture; shape/form factor; size/scale; notable parts; wear/defects; readable text/logos. Camera behavior, background context, and usage speculation are prohibited.

\noindent\textbf{Stage~2: Temporal-aware Captioning.}
Conditioned on the Stage~1 caption and a larger frame set (e.g., 24), it generates a fluent caption that integrates \emph{3–5} key appearance details with verified dynamics: camera motion, human–object interactions, and object motion.

\begin{figure*}[!t]
  \centering
  \includegraphics[width=\textwidth]{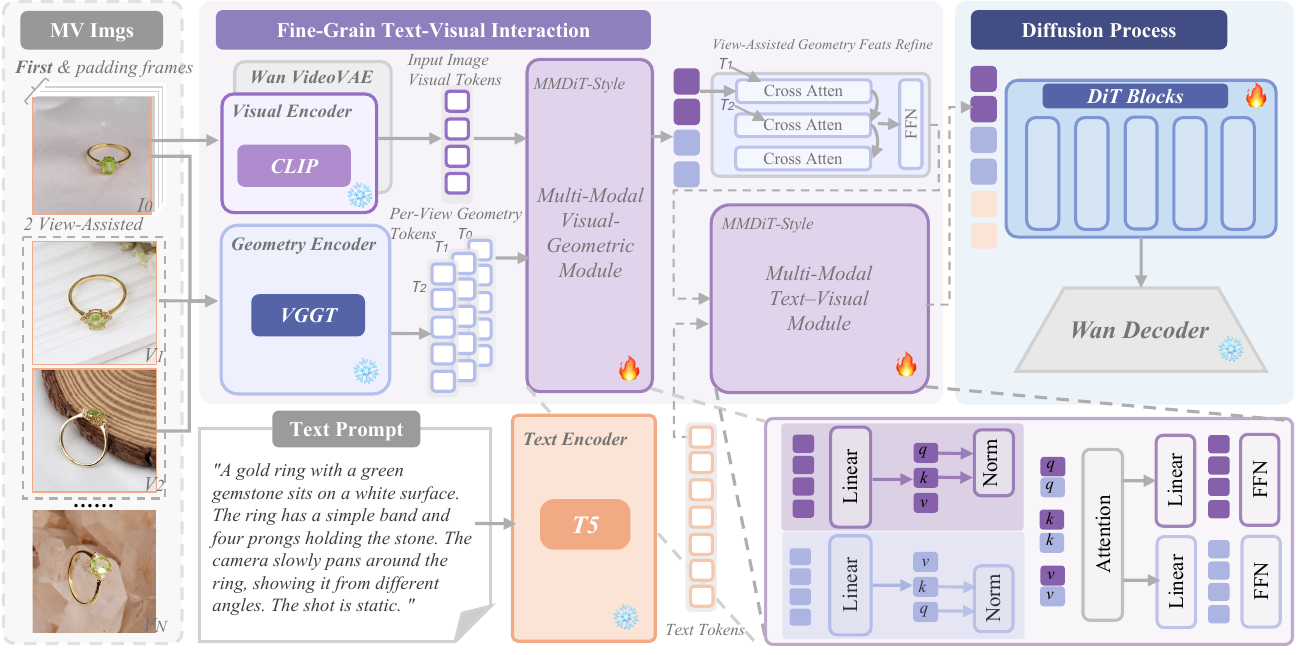}
  \vspace{-6mm}  
  \caption{\textbf{Overview of ConsID-Gen.} The model takes as input the first frame, two uncalibrated images, and a text instruction. Our Dual-Visual Encoder combines a Visual Encoder and a Geometry Encoder to extract visual-appearance and geometric representations. A unified multimodal interaction projector then fuses these features with the prompt to generate conditioning tokens for the DiT backbone.}
  \label{fig:architecture_overview}
  \vspace{-4mm}
\end{figure*}

\subsection{Synthetic Video Generation}

To enrich object‑ and viewpoint‑level diversity, we synthesize videos from MVImgNet2.0~\cite{han2024mvimgnet2} multi-view imagery. 
For each object, we select two representative views as start and end frames and extend video generator~\cite{zhang2025packing} into an interpolation variant. This produces smooth temporal sequences that preserve geometric consistency. 
Prompts are generated by Qwen2.5-VL~\cite{bai2025qwen2-5}, conditioned on the chosen start/end frames. 
Details provided in the Appendix~\ref{supp:synthetic_video_construction}.

\begin{figure}[b]
  \centering
  \includegraphics[width=\columnwidth]{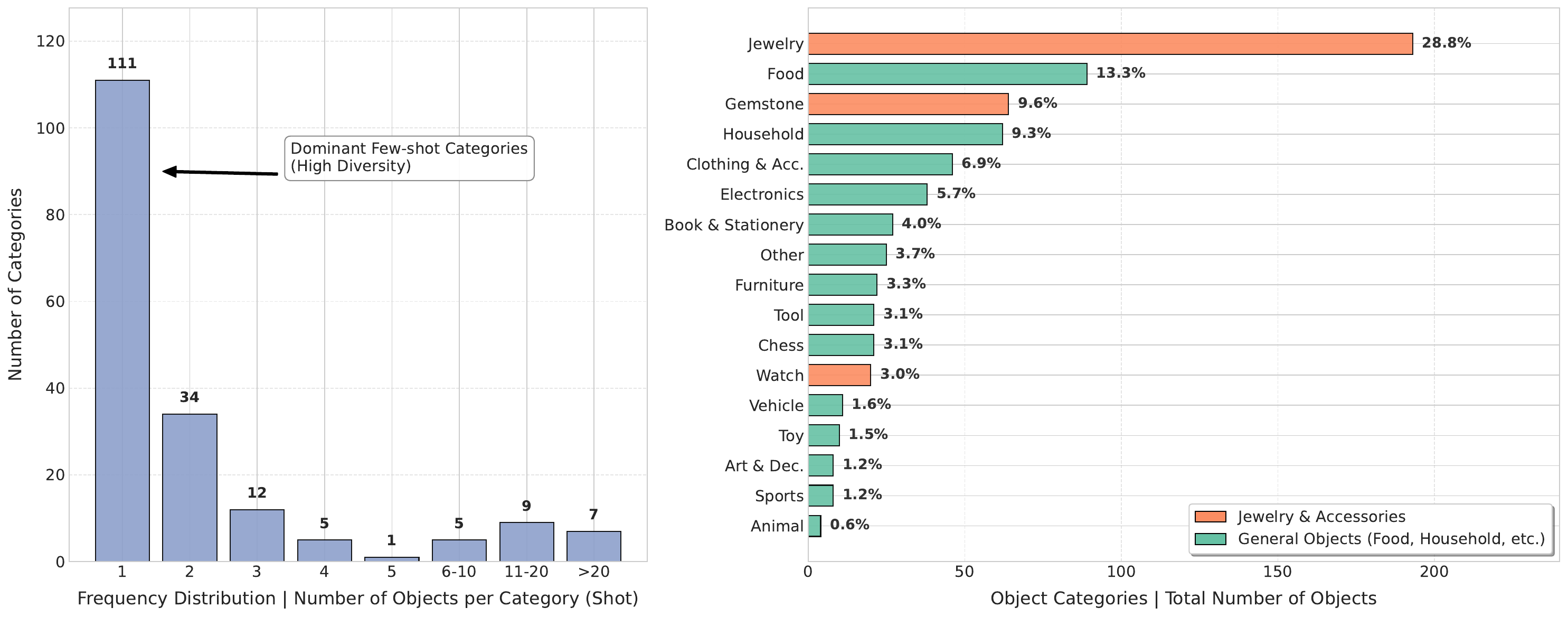}
  \caption{\textbf{Statistics of ConsIDVid-Bench.} Left: Frequency distribution of categories; Right: Object category breakdown.}
  \label{fig:data_distribution_statistics}
\end{figure}

\subsection{ConsIDVid-Bench}
\label{sec:considvid_benchmark}

To evaluate beyond scene-level semantics and frame-level fidelity, we introduce ConsIDVid-Bench, an object-centric benchmark for assessing identity preservation in \textit{I2V} generation. It aims to measure whether a generator maintains consistent object geometry and appearance under dynamic object or camera motion. By reformulating video evaluation as a multi-view consistency problem, ConsIDVid-Bench provides metrics sensitive to fine-grained appearance drift and geometric distortions over time.

\noindent\textbf{Task Definition.} Given an object-centric reference image $I_{\text{ref}}$ and a driving prompt $y$, the model is required to generate a temporally coherent video that maintains the object’s geometric and textural consistency while incorporating plausible object or camera motion.

\noindent\textbf{Evaluation Metrics.} 
We assess identity preservation with a comprehensive metric suite:
\textit{Chamfer Distance (CD)}, computed between 3D point sets reconstructed from the input and synthesized views, capturing global shape alignment and geometric stability over time; 
\textit{MEt3R}~\cite{asim2025met3r}, which applies DUSt3R~\cite{wang2024dust3r} to obtain dense pairwise reconstructions and measures cross-view feature similarity after projection; 
\textit{Video Similarity} (CLIP-based) to measure global realism and content consistency; 
and \textit{Object Similarity}, as illustrated in Fig.~\ref{fig:suppl_object_similarity_pipeline}, which uses DINO-based features on segmented objects to assess fine-grained identity preservation.
\section{Method}
\label{sec:methods}

In this section, we introduce ConsID-Gen, a view-assisted model for identity-preserving video diffusion.
Given a first frame $I_0$, two uncalibrated auxiliary images $\mathcal{V}=\{V_1,V_2\}$ of the same object, and a text instruction $y$, our goal is to synthesize a video $\mathcal{X}=\{X_t\}_{t=1}^T$ that maintains the object's identity throughout time.

\subsection{Model Architecture}
\label{sec:arch}

As illustrated in Fig.~\ref{fig:architecture_overview}, the model consists of a dual-visual encoder, a unified text–visual interaction projector, and a DiT backbone.
We build on Wan2.1 and explore strategies that strengthen identity preservation by jointly exploiting appearance and geometric cues.
Before detailing components, we motivate our dual-encoder formulation and the pre-alignment of visual and textual features.

\noindent\textbf{What hinders visual-conditioned I2V?}
Current \textit{I2V} pipelines derive dynamics from a first frame and a driving text prompt. The first frame is encoded by a pre-trained 2D encoder (e.g., CLIP~\cite{radford2021learning}) into semantic condition features that are fused with textual tokens via simple concatenation and a lightweight connector. 
While effective for high-level recognition, such 2D features under-represent fine-grained structure. 
During temporal synthesis the model tends to hallucinate missing spatial details, which leads to cumulative appearance drift and geometric distortions, particularly for rigid objects and under viewpoint changes. 
The core bottleneck is twofold: 2D observations are sparse and cross-modal alignment is weak, which together under-constrain the object geometry and its identity over time. 
Consistent with this diagnosis, Table~\ref{tab:t2v_i2v_comparison} shows that single-stream \textit{T2V} models, which do not require alignment between sparse visual and textual representations, tend to achieve stronger identity consistency.

We stabilize identity-preserving \textit{I2V} by (i) anchoring object shape and appearance with unposed multi-view reference imagery and (ii) introducing a dual-path visual representation that couples a semantic 2D encoder $E_{\mathrm{2D}}$ with a geometry-aware encoder $E_{\mathrm{geo}}$ trained to recover structural cues. 
The first frame, augmented with $\mathcal{V}$, provides local constraints on appearance and geometry, while the text prompt $y$ supplies global control over scene dynamics. 
A dedicated connector $g_{\phi}$ aligns and fuses semantic and geometric features with textual tokens, which mitigates modality misalignment and yields unified conditioning tokens for the DiT backbone $f_{\theta}$. 
We next detail these core components of the design.

\begin{table}[t]
    \footnotesize
    \centering
    \caption{Comparison of T2V and I2V models on identity preservation using automatic VBench metrics.}
    \label{tab:t2v_i2v_comparison}
    \vspace{-2mm}
    \setlength{\tabcolsep}{6pt}
    \resizebox{\linewidth}{!}{%
    \begin{tabular}{lcccc}
        \toprule
        & \multicolumn{4}{c}{Auto. Metrics} \\
        \cmidrule(lr){2-5}
        Method & T2V-S & T2V-B & I2V-S & I2V-B \\
        \midrule
        CogVideoX1.5-T2V~\cite{yang2024cogvideox}   & 95.77 & 96.31 & --    & --    \\
        CogVideoX1.5-I2V~\cite{yang2024cogvideox}   & 91.30 & 93.01 & 91.52 & 96.29 \\
        Wan2.1-T2V~\cite{wan2025wan}                & 96.72 & 97.10 & --    & --    \\
        Wan2.1-I2V~\cite{wan2025wan}                & 91.84 & 93.56 & 96.60 & 97.23 \\
        \bottomrule
    \end{tabular}
    }
    \vspace{-4mm}
\end{table}

\subsection{Dual-Visual Encoder}
Our model employs a dual-visual encoder composed of a 2D encoder \(E_{\text{2D}}\) and a geometry encoder \(E_{\text{geo}}\).

\noindent\textbf{2D Visual Encoder.} 
We use a CLIP-style image encoder \(E_{\text{2D}}\) to extract semantic appearance tokens from the first frame:
\[
F_{\text{2D}} = E_{\text{2D}}(I_0),
\qquad
F_{\text{2D}} \in \mathbb{R}^{\left\lfloor H/p_{\text{2D}}\right\rfloor \times \left\lfloor W/p_{\text{2D}}\right\rfloor \times d_{\text{2D}}},
\]
where \(H \times W\) is the image resolution, \(p_{\text{2D}}\) is the patch size, and \(d_{\text{2D}}\) is the feature dimension. These tokens provide high-level appearance priors for subsequent fusion.

\noindent\textbf{Geometric Encoder.} 
To complement semantic cues with geometry structure, we use VGGT~\cite{wang2025vggt} as the geometry backbone \(E_{\text{geo}}\). 
Given unposed auxiliary views \(\tilde{\mathcal{V}}=\{I_0, V_1, V_2\}\), each image is patchified and processed with alternating frame-wise and global self-attention to get dense geometry-aware tokens:
\[
F_{\text{geo}} = E_{\text{geo}}(\tilde{\mathcal{V}}),
\qquad
F_{\text{geo}} \in \mathbb{R}^{3 \times \left\lfloor H/p_{\text{geo}}\right\rfloor \times \left\lfloor W/p_{\text{geo}}\right\rfloor \times d_{\text{geo}}},
\]
where \(p_{\text{geo}}\) and \(d_{\text{geo}}\) denote the patch size and feature width of the geometry encoder, respectively. We retain the dense structural tokens for fusion with \(F_{\text{2D}}\).

\subsection{Multi-visual-text interaction}

After extracting semantic tokens \(F_{\text{2D}}\) and geometry-aware tokens \(F_{\text{geo}}\), we introduce a connector \(g_{\phi}\) that bridges modality gaps. 
It consists of a \emph{Multi-Modal Visual-Geometric module} that injects structural cues from \(F_{\text{geo}}\) into appearance tokens \(F_{\text{2D}}\), and a \emph{Multi-Modal Text–Visual module} that aligns the fused visual representation with text tokens \(T\) for fine-grained interaction.

\noindent\textbf{Multi-Modal Visual-Geometric Module.} 
Motivated by the dual-stream architecture of the Multimodal Diffusion Transformer (MMDiT)~\cite{esser2024scaling, wu2025qwen}, which enables effective alignment between visual and textual modalities, we extend this paradigm to the visual–geometric domain to achieve joint modeling of semantic appearance and 3D structure. 
Specifically, the Multi-Modal Visual–Geometric Module (MVGM) fuses appearance tokens \(F_{\text{2D}}\) with geometry-aware tokens \(F_{\text{geo}}\) extracted from the first frame $I_0$ through a dual-stream attention mechanism, enabling bidirectional interaction between semantic and structural cues. 
Furthermore, geometric features from the two auxiliary views $\mathcal{V}$ are integrated via cross-attention with the MVGM outputs, injecting multi-view structural priors that reinforce spatial and geometric consistency.

\noindent\textbf{Multi-Modal Text–Visual Module.}
Building on the fused visual–geometric representation, the Multi-Modal Text–Visual Module (MTVM) further aligns vision and language within a dual-stream attention mechanism. In this stage, textual features dynamically modulate the visual stream, while visual representations provide complementary cues to the text.

\begin{table*}[t]
\centering
\footnotesize
\setlength{\tabcolsep}{3.5pt}
\captionsetup{justification=raggedright, singlelinecheck=false}
\caption{Quantitative comparison on the proprietary subset of ConsIDVid-Bench. 
We evaluate model performance using VBench-I2V suite, Video Similarity, Object Similarity, Chamfer Distance, and MEt3R metrics. 
\colorbox{bestgreen}{Best} and \colorbox{secondgreen}{second-best} scores are highlighted 
}
\label{tab:consid_proprietary_set_results}
\vspace{-2mm}
\begin{tabular}{lcccccc|cc|cc}
\toprule
Method &
\makecell{I2V\\Subject} &
\makecell{I2V\\Background} &
\makecell{Subject\\Consistency} &
\makecell{Background\\Consistency} &
\makecell{Motion\\Smoothness} &
\makecell{Temporal\\Flickering} &
\makecell{Video\\Similarity} & 
\makecell{Object\\Similarity} &
\makecell{Chamfer \\ Distance} & 
\makecell{MEt3R} \\
\midrule
Wan2.1-1.3B~\cite{wan2025wan} &  96.22 &  97.12 &  91.03 &  94.57 &  \cellcolor{secondgreen}99.33 &  \cellcolor{secondgreen}98.84 &  87.15 & 66.9 & 0.1064 &  \cellcolor{secondgreen}0.1401 \\
SkyReelv2~\cite{chen2025skyreelsv2} &  94.03 &  95.21 &  85.61 &  92.04 &  98.71 &  97.42  & 85.33 & 59.5 & 0.1107 & 0.2177\\
ConsistI2V~\cite{ren2024consisti2v} &  94.93 &  93.42 &  91.41 &  94.07 &  98.25 &  96.72 & 82.48 & 62.0 & 0.1429 & 0.1614\\
Wan2.2-5B~\cite{wan2025wan2.2} &  \cellcolor{secondgreen}96.85 &  \cellcolor{secondgreen}97.57 &  \cellcolor{secondgreen}91.99 &  \cellcolor{secondgreen}94.82 &  98.93 &  98.10 &  \cellcolor{bestgreen}88.69 & \cellcolor{secondgreen}68.6 & \cellcolor{secondgreen}0.0921 & 0.1826 \\
CogVideoX1.5-5B~\cite{yang2024cogvideox} &  91.69 &  96.31 &  90.03 &  93.14 &  98.47 & 97.70 & 84.14 & 60.1 & 0.1194 & 0.1518\\
HunyuanVideo~\cite{kong2024hunyuanvideo} &  95.24 &  96.15 &  90.40 &  93.27 &  98.38 &  97.55 & 86.59& 64.3 & 0.1017 & 0.2270\\
Wan2.1-14B~\cite{wan2025wan} &  96.14 &  96.86 &  90.37 &  94.14 &  98.89 &  98.05 & 87.33 & 67.9 &  \cellcolor{bestgreen}0.0866 & 0.1572 \\
\midrule
\textbf{ConsID-Gen} &  \cellcolor{bestgreen}98.31 &  \cellcolor{bestgreen}98.66 &  \cellcolor{bestgreen}95.30 &  \cellcolor{bestgreen}96.10 &  \cellcolor{bestgreen}99.52 &  \cellcolor{bestgreen}99.24  &  \cellcolor{secondgreen}88.65 & \cellcolor{bestgreen}69.2 & 0.0996 &  \cellcolor{bestgreen}0.0978\\
\bottomrule
\end{tabular}
\end{table*}

\begin{table*}[t]
\centering
\footnotesize
\setlength{\tabcolsep}{3.5pt}
\captionsetup{justification=raggedright, singlelinecheck=false}
\caption{
Quantitative comparison on the public subset of ConsIDVid-Bench.  
We evaluate model performance using VBench-I2V suite, Video Similarity, Object Similarity, Chamfer Distance, and MEt3R metrics. 
\colorbox{bestgreen}{Best} and \colorbox{secondgreen}{second-best} scores are highlighted. 
}
\label{tab:consid_public_set_results}
\vspace{-2mm}
\begin{tabular}{lcccccc|cc|cc}
\toprule
Method &
\makecell{I2V\\Subject} &
\makecell{I2V\\Background} &
\makecell{Subject\\Consistency} &
\makecell{Background\\Consistency} &
\makecell{Motion\\Smoothness} &
\makecell{Temporal\\Flickering} &
\makecell{Video\\Similarity} & 
\makecell{Object\\Similarity} & 
\makecell{Chamfer \\ Distance} & 
\makecell{MEt3R} \\
\midrule
Wan2.1-1.3B~\cite{wan2025wan} & 97.34 & 97.71 & 92.86 & 94.09 & \cellcolor{bestgreen}99.32 & \cellcolor{bestgreen}98.48  & 83.37 & 69.1& 0.1503 & 0.1324 \\
SkyReelv2~\cite{chen2025skyreelsv2} & 96.59 & 97.00 & 91.67 & 93.23 & 99.16 & 97.92 & 84.80 & 68.0 & 0.1500 & 0.1526 \\
ConsistI2V~\cite{ren2024consisti2v} & 95.38 & 92.25 & 91.98 & 93.32 & 97.67 & 95.48 & 79.22 & 62.4 & 0.1700 & 0.1601 \\
Wan2.2-5B~\cite{wan2025wan2.2} & \cellcolor{bestgreen}98.47 & \cellcolor{bestgreen}98.64 & 94.02 & 94.39 & 98.85 & 97.47 & \cellcolor{secondgreen}84.81 & 71.6 & 0.1386 & 0.1591 \\
CogVideoX1.5-5B~\cite{yang2024cogvideox} & 91.26 & 96.14 & 90.58 & 92.05 & 98.91 & 97.84 & 80.26 & 61.5 & 0.1589 & 0.1409 \\ 
HunyuanVideo~\cite{kong2024hunyuanvideo} & 96.66 & 96.88 & 92.16 & 93.20 & 98.36 & 97.20 & 83.00 & 67.4 & \cellcolor{secondgreen}0.1377 & 0.2126 \\
Wan2.1-14B~\cite{wan2025wan} & \cellcolor{secondgreen}98.29 & \cellcolor{secondgreen}98.49 & \cellcolor{bestgreen}94.90 & 94.74 & 99.15 & 98.16 & 84.45 & \cellcolor{bestgreen}72.2 & 0.1322 & \cellcolor{bestgreen}0.0961 \\
\midrule
\textbf{ConsID-Gen} & 98.14 & \cellcolor{secondgreen}98.49 & \cellcolor{secondgreen}94.81 & \cellcolor{bestgreen}95.19 & \cellcolor{secondgreen}99.22 & \cellcolor{secondgreen}98.33 & \cellcolor{bestgreen}84.95 & \cellcolor{secondgreen}71.8 & \cellcolor{bestgreen}0.1277 & \cellcolor{secondgreen}0.1321 \\
\bottomrule
\end{tabular}
\end{table*}

\section{Experiments}
\label{sec:experiments}

In this section, we conduct comprehensive qualitative and quantitative evaluations of popular \textit{I2V} generators and our proposed ConsID-Gen to assess their capability for identity-preserving video generation.

\subsection{Experimental Settings}

\noindent\textbf{Implementation Details.}
We build our model upon Wan2.1-Fun-1.3B-InP~\cite{wan2025wan}, which generates 81-frame video clips at a $832\times480$ resolution. For training, we employ the Adam optimizer with a learning rate of $10^{-4}$. We use a per-GPU batch size of $1$ with gradient accumulation over $4$ steps (effective batch size $4$). The model is trained for 33K steps.
All experiments are conducted on NVIDIA A100 (80GB) GPUs. During inference, we utilize $50$ sampling steps and a classifier-free guidance (CFG) scale of $5$.

\noindent\textbf{Evaluation Metrics.} 
To evaluate identity consistency and temporally coherent dynamics, we adopt established metrics from the VBench-I2V~\cite{huang2024vbench++} suite: Subject Consistency, Background Consistency, Motion Smoothness, and Temporal Flickering. To further assess the fidelity of identity preservation, we employ geometry-aware metrics, including MEt3R, Chamfer Distance, and Video Similarity.

\noindent\textbf{Evaluation Datasets.} 
We conduct quantitative evaluation using our proposed \textbf{ConsIDVid-Bench}. As detailed in Section~\ref{sec:considvid_benchmark}, this benchmark is specifically designed to assess identity preservation and comprises two subsets: the proprietary subset (241 videos), which consists of the videos depicting the product of popular e-commerce listings,
and the public subset (370 videos), built from existing object-centric datasets and synthetic videos. 

\subsection{Quantitative Evaluations}

\noindent\textbf{Results on the proprietary Subset.}
Table~\ref{tab:consid_proprietary_set_results} presents the evaluation results on our proprietary subset. 
ConsID-Gen achieves state-of-the-art (SOTA) performance across the VBench-I2V suite. 
Compared to the strong Wan2.2~\cite{wan2025wan2.2}, ConsID-Gen demonstrates higher identity fidelity, achieving a 3.6\% higher score in Subject Consistency. Notably, ConsID-Gen yields a substantially lower score in the geometry-aware MEt3R~\cite{asim2025met3r} metric, demonstrating superior multi-view consistency. 
While Wan2.2 leads slightly in Video Similarity and Wan2.1-14B achieves the best Chamfer Distance, ConsID-Gen remains highly competitive.

\noindent\textbf{Results on the public Subset of ConsIDVid-Bench.}
As shown in Table~\ref{tab:consid_public_set_results}, ConsID-Gen demonstrates highly competitive performance on the public subset. 
Notably, ConsID-Gen achieves superior performance in geometric and fidelity metrics, achieving the top scores for both Chamfer Distance and Video Similarity. 
However, we observe that ConsID-Gen yields suboptimal results for I2V Subject and I2V Background compared to other methods~\cite{wan2025wan2.2}. 
We partially attribute this to a qualitative artifact: when the input contains distracting structures (\textit{e.g., grid paper}), our generated videos occasionally suffer from degradation or collapse, 
a phenomenon that is analyzed in detail in the Appendix~\ref{suppl:failure_cases}.

\begin{figure*}[t]
    \centering
    \vspace{-2mm}
    \includegraphics[width=\textwidth]{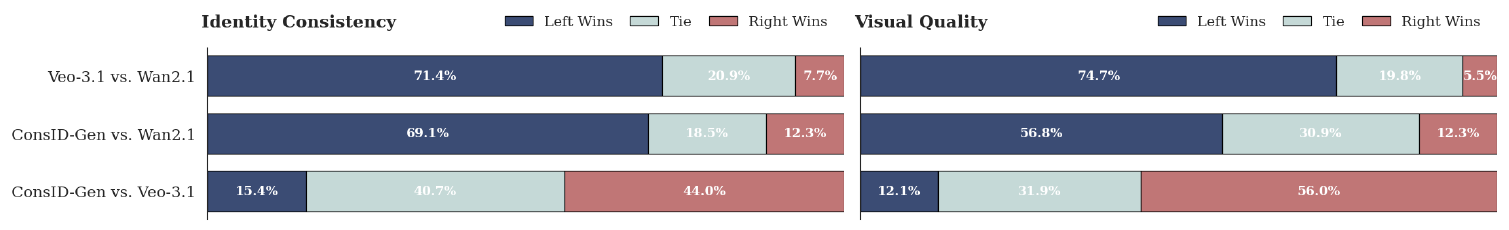}
    \caption{\textbf{Human Evaluation} results for Identity Consistency (left) and Visual Quality (right).}
    \label{fig:user_study}
    \vspace{-2mm}
\end{figure*}

\noindent\textbf{Human Preference Evaluation.} 
We conducted a side-by-side user study to benchmark ConsID-Gen against the open-source Wan2.1~\cite{wan2025wan} and proprietary Veo-3.1~\cite{veo3}. 
Participants were presented with randomized video pairs and asked to express their preference (better or tie) regarding \textit{Identity Consistency} and \textit{Visual Quality}. 
As shown in Figure~\ref{fig:user_study}, our method consistently outperforms Wan2.1 across both metrics. Compared to Veo-3.1, we achieve comparable results in identity consistency.

\begin{figure*}[t]
    \centering
    \includegraphics[width=\textwidth]{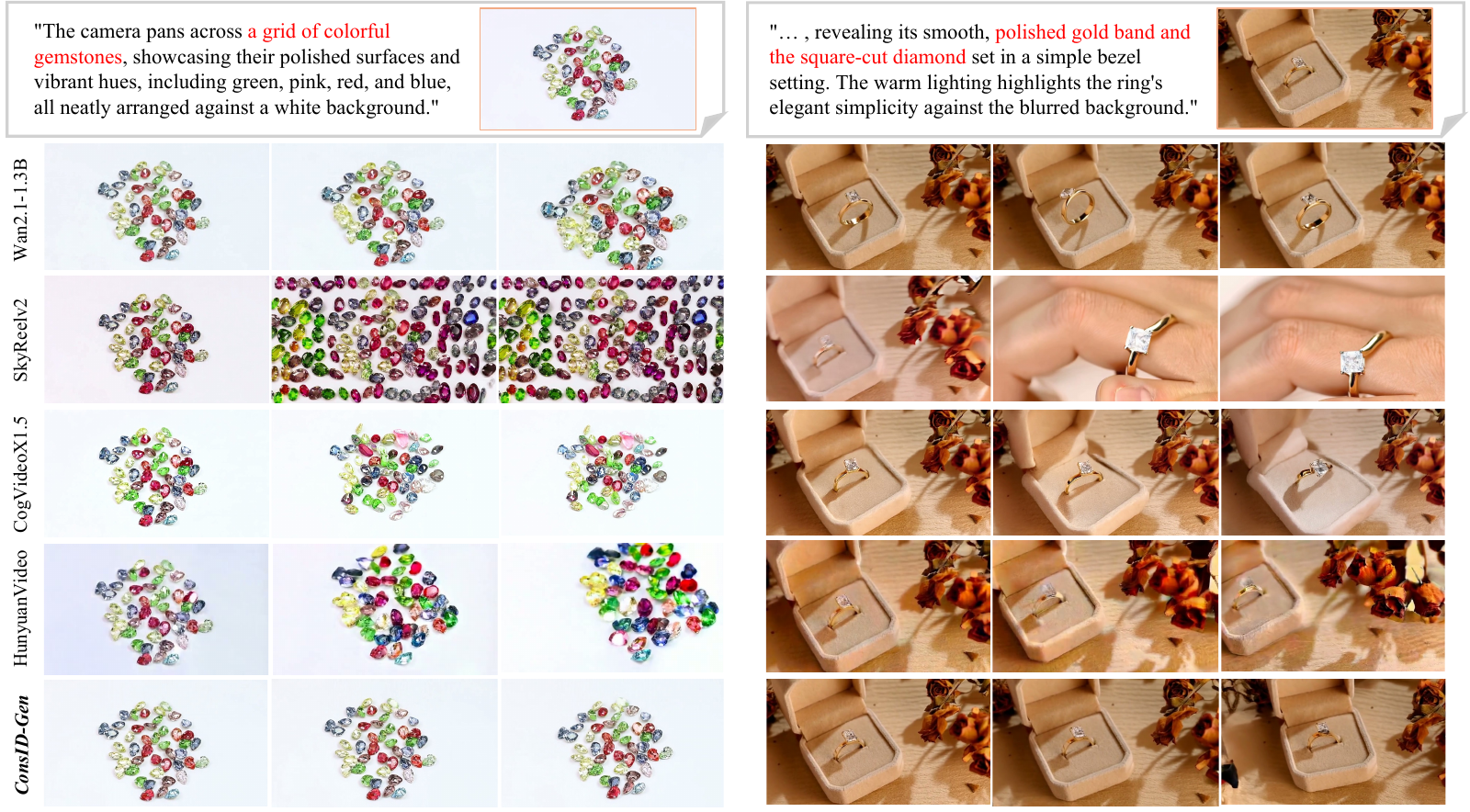}
    \caption{\textbf{Qualitative comparison with popular \textit{I2V} methods.} ConsID-Gen maintains the identity and geometry of objects in challenging scenarios. Compared to existing methods, our results demonstrate superior geometric fidelity and temporal coherence.}
    \label{fig:qualitative_comparison}
\end{figure*}

\subsection{Qualitative Evaluations}

Figure~\ref{fig:qualitative_comparison} presents the qualitative comparisons between ConsID-Gen and existing methods. As illustrated, ConsID-Gen generates videos with strong identity preservation, avoiding issues of appearance drift or geometric collapse. In contrast, videos produced by existing popular methods exhibit noticeable inconsistencies and temporal artifacts. 
For instance, in the "gemstone" example (left), methods~\cite{kong2024hunyuanvideo, chen2025skyreelsv2} suffer from object jitter and severe scene changes. 
In the "ring" example (right), other methods~\cite{yang2024cogvideox} fail to maintain geometric integrity, causing the subject to visibly deform.

\begin{figure}[t]
  \centering
  \includegraphics[width=\columnwidth]{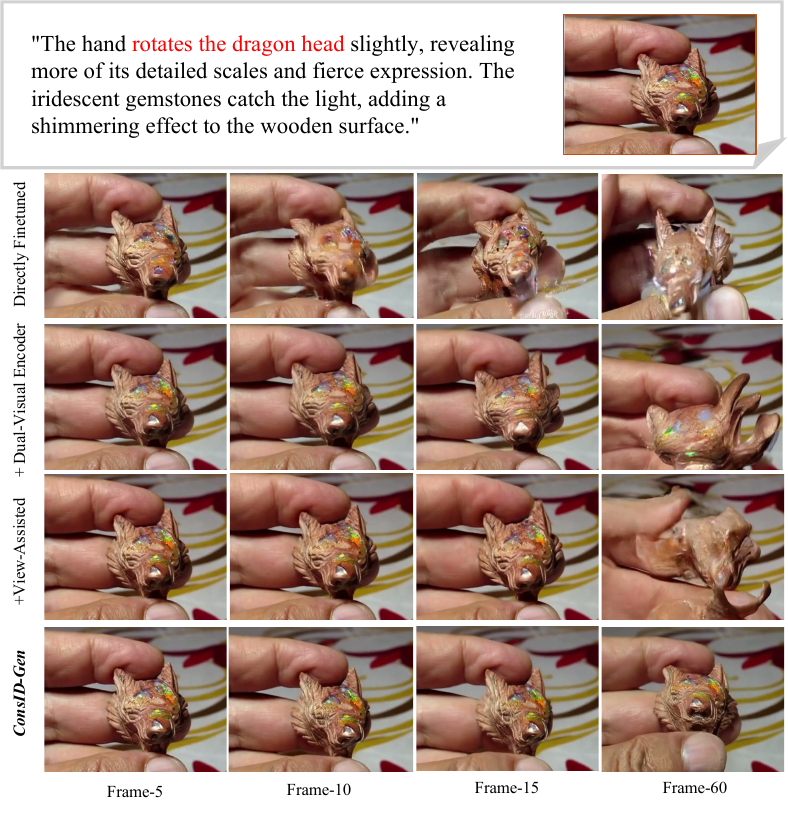}
  \caption{\textbf{Qualitative results of the ablation study.} ConsID-Gen maintains consistent identity across longer temporal spans.}
  \label{fig:ablation_components_fig}
\end{figure}

\subsection{Ablation Studies}

\noindent \textbf{Effect of Key Components.} 
We conduct ablation studies to validate our key architectural components. Due to computational resource constraints, these ablated models were finetuned on 50\% of the training data and evaluated on a randomly sampled 60-video subset.
The results in Table~\ref{tab:ablation_components_main} reveal a clear progression: we find that the geometry encoder (“+ Geo Enc.") in isolation provides no significant benefits over the baseline. However, further adding multiple unposed, view-assisted images (“+ View-Asst.") yields a clear improvement.
These findings are further supported by what is illustrated in Figure~\ref{fig:ablation_components_fig}, where we observe that direct finetuning Wan2.1 still leads to noticeable identity shift in the early frames of the generated videos. Incorporating the geometry encoder and view-assisted images is able to alleviate this issue to some extent. 
However, our full ConsID-Gen model that fuses text and visual cues ensures long-range identity stability.

\begin{table}[t]
\centering
\footnotesize
\setlength{\tabcolsep}{1.5pt}
\renewcommand{\arraystretch}{1.0}
\vspace{-3mm}
\caption{Quantitative ablation of key components. We evaluate model performance using VBench metrics and Video Similarity.}
\label{tab:ablation_components_main}
\vspace{-2mm}
    \resizebox{\linewidth}{!}{%
\begin{tabular}{lcccc|c}
\toprule
\textbf{Method} &
I2V-Subj &
I2V-Back &
Subj-Cons. &
Back-Cons. &
Video-Sim. \\
\midrule
Baseline & 96.30 & 97.16 & 90.83 & 94.97 & 87.75 \\
+ Geo Enc. & 96.29 & 97.37 & 89.65 & 93.44 & 86.19 \\
+ View-Asst. & 96.97 & 97.85 & 91.87 & 94.33 & 87.35 \\
\midrule
\textbf{ConsID-Gen} & \textbf{98.48} & \textbf{98.85} & \textbf{95.13} & \textbf{96.20} & \textbf{88.25} \\
\bottomrule
\end{tabular}}
\end{table}


\begin{table}[t]
\centering
\scriptsize
\setlength{\tabcolsep}{3pt} %
\vspace{-2mm}
\caption{Quantitative ablation of dataset effectiveness. We evaluate model performance using  VBench-I2V suite.}
\label{tab:ablation_effect_of_data}
\vspace{-2mm}
\begin{tabular}{lcccccc}
    \toprule
    Method & I2V-Subj & I2V-Back & Subj-Cons. & Back-Cons. & Motion & Temp \\    
    \hline
    Wan2.2-5B    & 96.85 &  97.57 &  91.99 &  94.82 &  98.93 &  98.10 \\
    Wan2.2-5B-FT & 97.61 &  98.17 &  91.22 &  94.64 &  99.21 &  98.37 \\
    \hline
    \textbf{ConsID-Gen} & \textbf{98.31} &  \textbf{98.66} &  \textbf{95.30} &  \textbf{96.10} &  \textbf{99.52} &  \textbf{99.24}\\
    \bottomrule
\end{tabular}
\vspace{-2mm}
\end{table}

\noindent \textbf{Effect of Datasets.} 
To investigate the impact of the data, we fine-tuned Wan2.2-5B via LoRA (rank 64). As shown in Table~\ref{tab:ablation_effect_of_data}, the resulting Wan2.2-5B-FT showed limited improvement compared to ConsID-Gen, which achieved significantly better results. 
This underscores that our architectural design is the key factor behind the performance boost.

\section{Conclusion}
\label{sec:conclusion}

In this paper, we discuss the preservation of identity in \textit{I2V} from both the data and the model perspectives.  On the data side, we curate ConsIDVid, a large-scale object-centric dataset, and introduce ConsIDVid-Bench, which reframes evaluation as multi-view consistency to capture precise geometric and appearance drift. On the model side, we propose ConsID-Gen, a View-Assisted Video Generation framework that augments the first frame with unposed auxiliary views and performs fine-grained pre-alignment via dual-stream visual–geometric fusion and a text–visual connector. 
Across proprietary and public subsets of ConsIDVid-Bench, our model consistently exceeds baselines with stronger identity fidelity under challenging real-world scenes. 

\section*{Acknowledgments}
We thank Siyuan Yang for the discussions. We also thank Yixin Chen, Zhe Dong, Kuan-Ru Huang, Yanjia Huang, Zhaoming Xu, Jongze Yu, Zihao Zhu, and Yushen Zuo for their assistance with the user studies.

{\small
\bibliographystyle{ieeenat_fullname}
\bibliography{main}}

\clearpage
\setcounter{page}{1}
\maketitlesupplementary
\setcounter{section}{0}

\section{ConsIDVid Dataset: Additional Details}

ConsIDVid is primarily built upon real-world, object-centric videos collected from public sources~\cite{reizenstein21co3d, wu2023omniobject3d} and is additionally supplemented by proprietary datasets.

\subsection{Synthetic Video Construction}
\label{supp:synthetic_video_construction}

\noindent\textbf{Synthetic Video Generation.} 
To significantly enhance the dataset's diversity and coverage, we generate synthetic videos utilizing FramePack~\cite{zhang2025packing}, a video generator built upon HunyuanVideo~\cite{kong2024hunyuanvideo}.
Given that the single-image conditioning used in the standard FramePack pipeline offers limited visual guidance, we extend the framework to support synthesis by conditioning on start and end keyframes.

\noindent\textbf{Controlled Keyframe Selection Strategy.} 
For the synthetic samples derived from MVImgNet2.0~\cite{han2024mvimgnet2}, we consciously avoid directly stitching its complete multi-view image sequences. 
These sequences frequently exhibit rapid camera motion or contain multiple full rotations around the object, which leads to excessive viewpoint shifts. 
Instead, we employ a controlled strategy where the first frame of each sequence is designated as the starting keyframe, and the ending keyframe is selected from indices 4 through 8 based on the LAION aesthetic predictor~\cite{schuhmann2022laion}.

\subsection{Hierarchical Video Captioning}
In Section~\ref{hierarchical_video_captioning}, 
we adopt a Hierarchical Video Captioning strategy to construct video captions in a structured manner. This process involves generating captions by two distinct levels. The detailed instruction templates used for both levels of caption generation are illustrated in Figure~\ref{box:template-appearance-caption} and Figure~\ref{box:template-temporal-caption}, respectively.

\begin{table*}[!b]
\captionsetup{skip=6pt}
\caption{Comparison of existing domain-specific video generation datasets and our ConsIDVid.}
\label{tab:dataset_comparison}
\centering
\small
\setlength{\tabcolsep}{6pt}
\begin{tabular}{l c c c c c c c c}
\toprule
Dataset & Year & Scenario & \#Videos & Avg. Len (s) & Dur. (h) & Resolution & Caption & Motion Type \\
\midrule
UCF-101~\cite{soomro2012ucf101}            & 2012 & Human        & 13.3K & 7.2  & 26.7 & 240p      & Short        & Text \\
MSP-Avatar~\cite{sadoughi2015msp}          & 2015 & Human        & 74    & --   & 3    & 1080p     & N/A          & Landmark, Pose \\
Taichi-HD~\cite{siarohin2019first}         & 2019 & Human        & 3K    & --   & --   & 256p      & Short        & Text \\
TikTok-v4~\cite{chang2023magicpose}        & 2023 & Human        & 350   & --   & 1    & --        & N/A          & Skeleton \\
SkyTimelapse~\cite{xiong2018learning}      & 2018 & Sky          & 35K   & --   & --   & 360p      & N/A          & -- \\
FaceForensics++~\cite{rossler2019faceforensics++} 
                                           & 2019 & Face         & 1K    & --   & --   & Diverse   & N/A          & -- \\
CelebV-HQ~\cite{zhu2022celebv}             & 2022 & Portrait     & 35K   & 6.6  & 68   & 512p      & N/A          & -- \\
ChronoMagic~\cite{yuan2024chronomagic}     & 2024 & Metamorphic  & 2K    & 11.4 & 7    & Diverse   & Long         & Text \\
\midrule
\textbf{ConsIDVid}                         & 2025 & Rigid Object & 44.3K & 8.4  & 104  & Diverse   & Hierarchical & Text, Images \\
\bottomrule
\end{tabular}
\end{table*}

\subsection{Comparison with Existing Video Datasets}

Recent efforts~\cite{chen2024panda, wang2025koala, nan2024openvid} in video generation primarily focus on collecting large, general-purpose video datasets to train video generative models. However, domain-specific video datasets remain limited in both scale and diversity. As shown in Table \ref{tab:dataset_comparison}, many existing domain-focused resources are mostly human-centric (e.g., UCF-101~\cite{soomro2012ucf101}, Taichi-HD~\cite{siarohin2019first}, FaceForensics++~\cite{rossler2019faceforensics++}), making them inadequate for capturing fine-grained object identity or rigid-object motion patterns. While prior approaches like Track4Gen~\cite{jeong2025track4gen} relied on small and minimally curated appearance-preserving datasets, we introduce ConsIDVid. This large-scale, object-centric, identity-preserving video dataset, curated via a scalable pipeline, also includes an appearance-preserving benchmark for standardized evaluation of I2V models.

\section{ConsIDVid-Bench: Evaluation Metrics}

An ideal Image-to-Video (I2V) generator must not only align with the text prompt but, crucially, preserve visual fidelity throughout the temporal dynamics. 
Accurately quantifying appearance drift and geometric distortion is paramount for fine-grained video generation evaluation. 
Therefore, in our experiments, we utilize established evaluation metrics from VBench~\cite{huang2024vbench, huang2024vbench++} while introducing novel Multi-View Consistency metrics to rigorously measure view and object fidelity.

\begin{figure*}[!t]
  \captionsetup{skip=6pt}
  \centering
  \includegraphics[width=\textwidth]{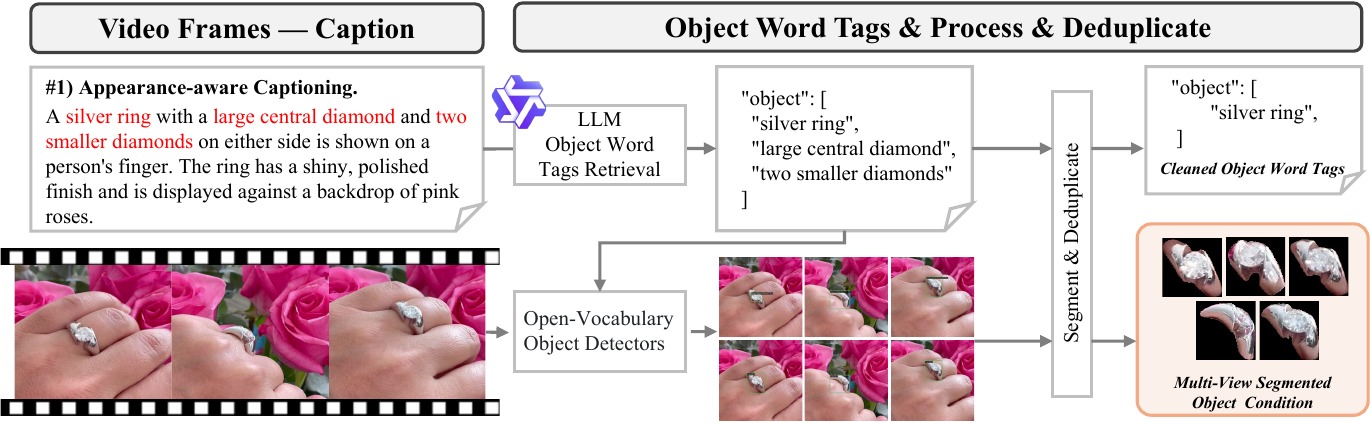}
  \caption{\textbf{Overview of object similarity pipeline.} It extracts clean multi-view object segments via caption-based word retrieval, open-vocabulary detection, segmentation, and de-duplication.}
  \label{fig:suppl_object_similarity_pipeline}
  \vspace{-1mm}
\end{figure*}

\subsection{VBench Metrics for I2V Evaluation}

VBench extends Text-to-Video (T2V) metrics to the I2V domain, focusing on semantic and temporal consistency.

\begin{itemize}
    \item \textbf{I2V Subject}: Cosine similarity between DINO~\cite{caron2021emerging} features of the input image and the generated frames, measuring the preservation of the subject from the input image within the generated video.
    
    \item \textbf{I2V Background}: DreamSim~\cite{fu2023dreamsim} feature similarity between the input image and generated frames, assessing the visual consistency of the scene/background.
    
    \item \textbf{Subject Consistency}: Average cosine similarity of DINO features across consecutive frames, evaluating subject appearance consistency throughout the video.

    \item \textbf{Background Consistency}: Average cosine similarity of CLIP~\cite{radford2021learning} features across consecutive frames, measuring the temporal consistency of the background scene.

    \item \textbf{Motion Smoothness}: Motion prior score derived from a video frame interpolation model~\cite{li2023amt}, assessing whether the generated motion remains smooth.

    \item \textbf{Temporal Flickering}: Mean absolute difference between consecutive frames at the pixel level, detecting high-frequency artifacts and local temporal inconsistencies in the generated video.
\end{itemize}

\subsection{Multi-View Metrics for I2V Evaluation}

Instead of relying solely on single-frame image-to-video similarity, we further assess video consistency by sampling multi-view (multi-frame) observations from the ground-truth video. This approach allows for a more rigorous measurement of fine-grained identity preservation via the following proposed metrics:

\begin{itemize}
    \item \textbf{Video Similarity}: Average cosine similarity between CLIP features of the ground-truth and generated sampled frames. This metric quantifies overall video realism and content preservation.
    
    \item \textbf{Object Similarity}: Average cosine similarity between DINO features of the segmented objects in the reference images and the corresponding segments in the generated frames. For rigorous evaluation, multiple reference embeddings per object category are used, and missing objects receive a fixed low similarity penalty. This metric further assesses fine-grained object identity preservation.
\end{itemize}

\subsection{Geometric-aware Metrics for I2V Evaluation}

In the context of rigid-object centric I2V generation, the synthesized video can be viewed as a multi/cross-view image sequence derived from a single input image. Crucially, these generated images must exhibit 3D geometric consistency to form a coherent object representation over time. While an I2V generator may produce frames that diverge from the ground-truth, we fundamentally require them to be geometrically consistent with each other.

To address the limitations inherent in ground-truth-dependent geometric evaluation, our key idea is to measure geometric consistency via self-consistency in 3D between the generated multi-view videos. We quantify the geometric fidelity of the I2V output using the following metrics:

\begin{itemize}
    \item \textbf{Chamfer Distance (CD)}: To evaluate this property, we reconstruct 3D point clouds from sampled frames of the generated videos using VGGT~\cite{wang2025vggt}, followed by point filtering and rigid alignment via ICP (Iterative Closest Point). 
    Our ground-truth point cloud is similarly generated from the true video frames using the same VGGT pipeline.
    Following prior work on multi-view consistency, we then measure the bidirectional geometric discrepancy between two reconstructed point clouds. 
    This metric captures global shape alignment while penalizing geometric drift or deformation across the synthesized views.

    \item \textbf{MEt3R}~\cite{asim2025met3r}: This metric evaluates view consistency by employing DUSt3R~\cite{wang2024dust3r} to obtain dense 3D reconstructions from image pairs. 
    It measures consistency by projecting DINO + FeatUp~\cite{fu2024featup} features from one view to the other using the reconstructed geometry and calculating the feature similarity among the resulting views. 
    This provides a reliable measure of geometric self-consistency for multi-view coherence in generated images.
\end{itemize}

\begin{table*}[!t]
\small
\centering
\caption{Quantitative comparison of model performance on ConsIDVid-Bench under two penalty settings ($\text{penalty}=0.1$ and $0.5$), evaluated by object similarity. Inference latency is measured on a single NVIDIA A100 GPU. \colorbox{bestgreen}{Best} and \colorbox{secondgreen}{Second-best} scores are highlighted.}
\label{tab:suppl_object_similarity}
\begin{tabular}{l|c|c|cc|cc}
\toprule
\multirow{2}{*}{Model} & \multirow{2}{*}{Params} & \multirow{2}{*}{Latency}
& \multicolumn{2}{c|}{Penalty = 0.1} 
& \multicolumn{2}{c}{Penalty = 0.5} \\
\cmidrule{4-7}
& & & Proprietary & Public & Proprietary & Public \\
\midrule
Wan2.1~\cite{wan2025wan} & 1.3B & \cellcolor{secondgreen}202 (s) & 66.9 & 69.1 & 67.1 & 69.6 \\
SkyReelv2~\cite{chen2025skyreelsv2} & 1.3B & 393 (s) & 59.5 & 68.0 & 60.0 & 68.5 \\
\midrule
ConsistI2V~\cite{ren2024consisti2v} & 5.2B & -- & 62.0 & 62.4 & 62.7 & 63.1 \\
Wan2.2~\cite{wan2025wan2.2} & 5B & 359 (s) &
\cellcolor{secondgreen}68.6 & 71.6 &
\cellcolor{secondgreen}68.9 & 72.1 \\
CogVideoX1.5~\cite{yang2024cogvideox} & 5.2B & -- & 60.1 & 61.5 & 60.5 & 62.1 \\
HunyuanVideo~\cite{kong2024hunyuanvideo} & 13B & -- & 64.3 & 67.4 & 64.6 & 67.6 \\
Wan2.1~\cite{wan2025wan} & 14B & 970 (s) & 67.9 & \cellcolor{bestgreen}72.2 & 68.2 & \cellcolor{bestgreen}72.8 \\
\midrule
\textbf{ConsID-Gen} & 1.8B & \cellcolor{bestgreen}199 (s) &
\cellcolor{bestgreen}69.2 & \cellcolor{secondgreen}71.8 &
\cellcolor{bestgreen}69.9 & \cellcolor{secondgreen}72.3 \\
\bottomrule
\end{tabular}
\end{table*}

\section{Object Similarity Evaluation}
As illustrated in Figure~\ref{fig:suppl_object_similarity_pipeline}, we propose an object similarity evaluation designed to measure fine-grained appearance consistency across generated videos. 
Our method utilizes multi-view frames rather than relying solely on the first frame, as it is in the I2V Subject, thereby ensuring robustness against background variations. 
First, we employ a Large Language Model (LLM)~\cite{yang2025qwen3} for the first stage output of Hierarchical Video Captioning to retrieve object-related word tags. 
Next, these tags guide an open-vocabulary object detector~\cite{fu2025llmdet} to localize objects in 5 sampled keyframes, followed by segmentation~\cite{ravi2024sam}.
Finally, to ensure data reliability, we de-duplicate and consolidate these instances, yielding a reliable set of cleaned object word tags and their corresponding segmented visual counterparts for precise, object-level comparison.


\section{More Comparison Evaluations}

\subsection{Quantitative Evaluation on ConsIDVid-Bench}

\noindent\textbf{Object Similarity.} As shown in Table~\ref{tab:suppl_object_similarity}, ConsID-Gen achieves consistently strong performance across both evaluation settings. Under the default penalty of 0.1, where video frames with missing objects are assigned a similarity score of $0.1$, ConsID-Gen obtains the highest Object Similarity score on the proprietary set, surpassing all competing models of similar or larger scale. A similar trend can be seen when comparing the results for a higher penalty of $0.5$. This robust performance indicates that ConsID-Gen effectively maintains stable object visibility and identity fidelity throughout the generated video sequence.

\section{More Ablation Evaluations}

\subsection{Effective Evaluation on Video Captioning}

To assess our Hierarchical Video Captioning strategy, we compared its Stage 1 (Appearance-aware Captioning) against a normal captioning method. Unlike the normal method, which jointly reasons over appearance and temporal dynamics, our Stage 1 processes fewer frames at higher resolution, allowing the VLM~\cite{bai2025qwen2} to focus exclusively on fine-grained object details. Evaluating both on 50 proprietary videos by retrieving object word tags using an LLM~\cite{yang2025qwen3}, we found that Stage 1 yields richer and more precise appearance-centric tags, as presented in Table~\ref{tab:suppl_word_tag_normal_vs_stage1}.


\begin{table}[t]
\centering
\footnotesize
\caption{Effectiveness of Hierarchical Video Captioning. Comparison of object word tag retrieval between normal captioning and our Stage-1 appearance-aware captioning, demonstrating that the latter yields richer and more precise appearance tags.}
\label{tab:suppl_word_tag_normal_vs_stage1}
\begin{tabular}{l|c|c}
\toprule
Method & Avg. Objects / Video & Avg. Word Len. \\
\midrule
Normal Caption & 3.18 & 13.69 \\
Appearance-aware Caption & 3.20 & 14.44 \\
\bottomrule
\end{tabular}
\end{table}

\section{Miscellaneous Visualization Results}

\subsection{Effective Evaluation on Video Captioning}

\begin{figure*}[!t]
  \captionsetup{skip=6pt}
  \centering
  \includegraphics[width=\textwidth]{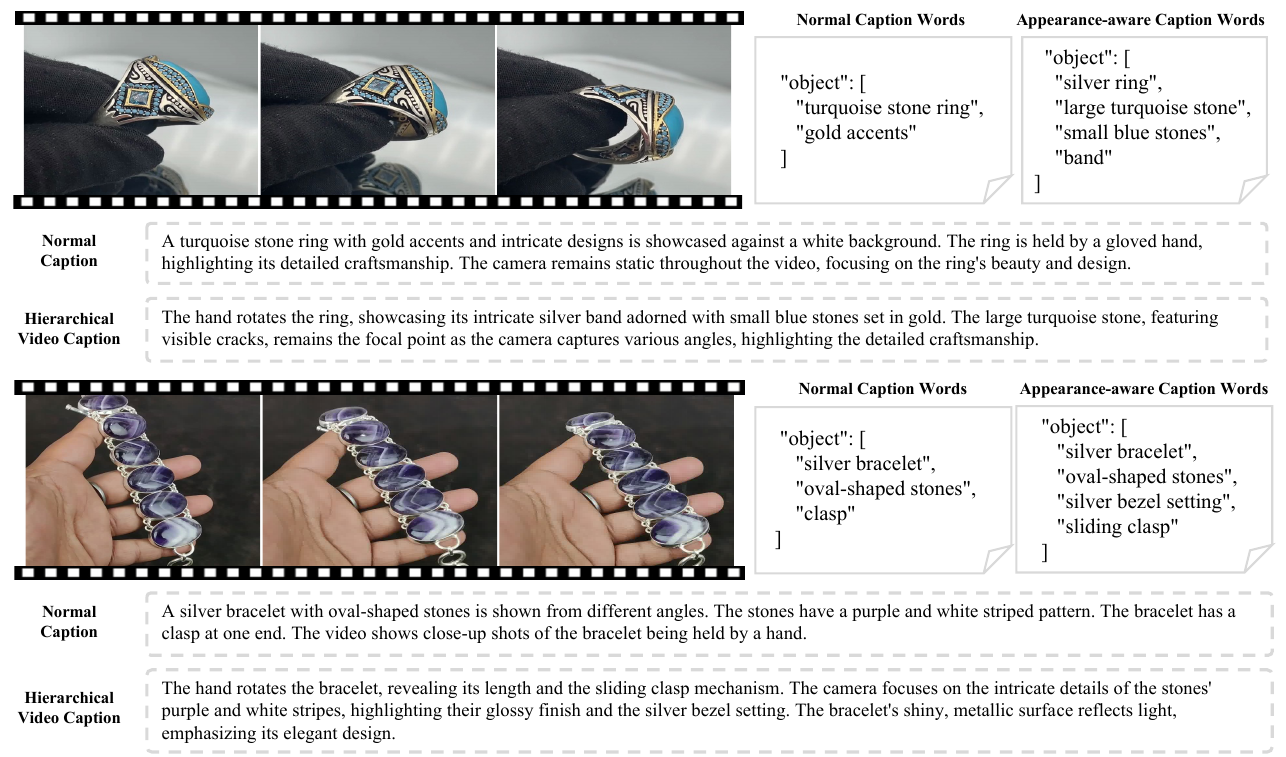}
  \caption{\textbf{Qualitative comparison between normal captioning and Hierarchical Video Captioning.} The right side displays the retrieved object word tags from the normal caption and the appearance-aware caption, while the lower section illustrates the captions generated by normal and hierarchical captioning.}
  \label{fig:suppl_caption_word_tags}
\end{figure*}

As illustrated in Figure~\ref{fig:suppl_caption_word_tags}, Stage-1 appearance-aware captioning produces richer and more precise appearance tags than normal captioning, capturing fine-grained object details such as material, stone texture, and setting structure. 
Moreover, the hierarchical design enables the model to first ground stable appearance semantics before generating temporal descriptions, yielding more comprehensive and accurate descriptions.

\subsection{Additional Comparison with Existing Methods}
\begin{figure*}[!t]
  \centering
  \includegraphics[width=0.98\textwidth]{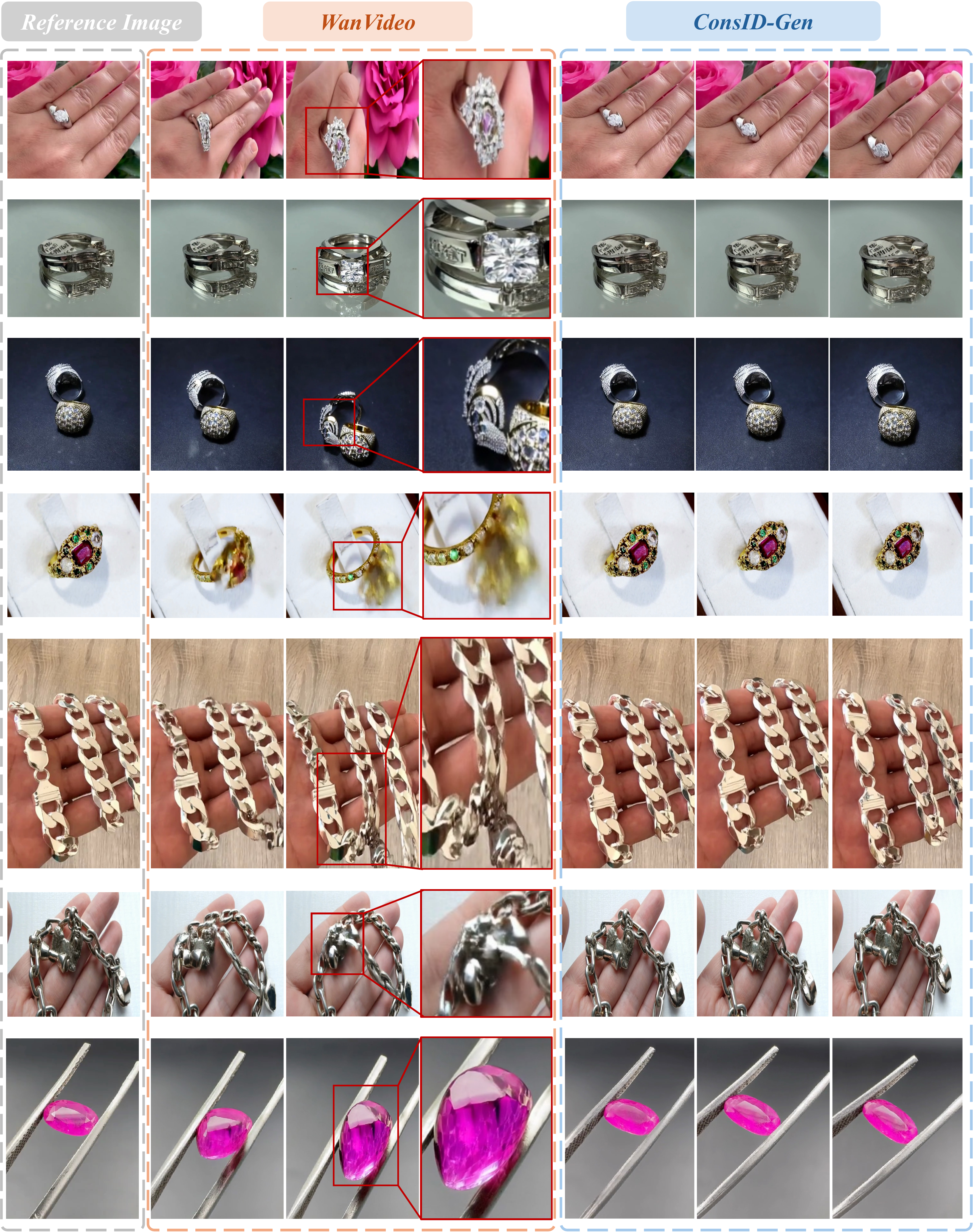}
  \caption{\textbf{Additional visual comparisons on the proprietary subset of ConsIDVid-Bench.}}
  \label{fig:suppl_more_comparison_on_properity_subset}
\end{figure*}

\begin{figure*}[!t]
  \centering
  \includegraphics[width=\textwidth]{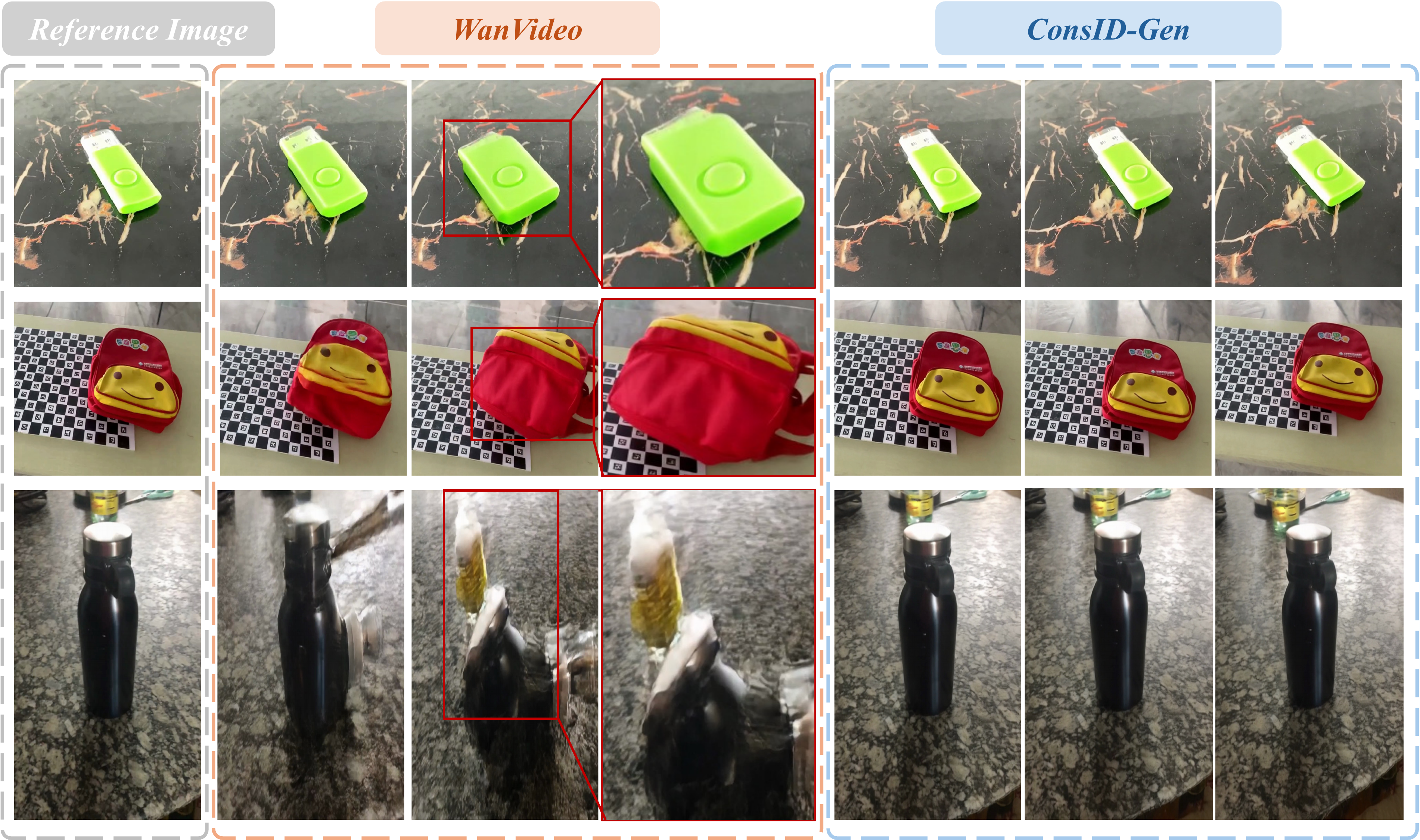}
  \caption{\textbf{Additional visual comparisons on the public subset of ConsIDVid-Bench.}}
  \label{fig:suppl_more_comparison_on_public_subset}
\end{figure*}

To complement the quantitative evaluations, we provide additional synthetic samples generated by existing methods. Figure~\ref{fig:suppl_more_comparison_on_properity_subset} and Figure~\ref{fig:suppl_more_comparison_on_public_subset} illustrate these samples on the proprietary and public subsets of ConsIDVid-Bench.

\subsection{Visualization of Failure Cases} 
\label{suppl:failure_cases}
\begin{figure}[H]
  \captionsetup{skip=6pt}
  \centering
  \includegraphics[width=\columnwidth]{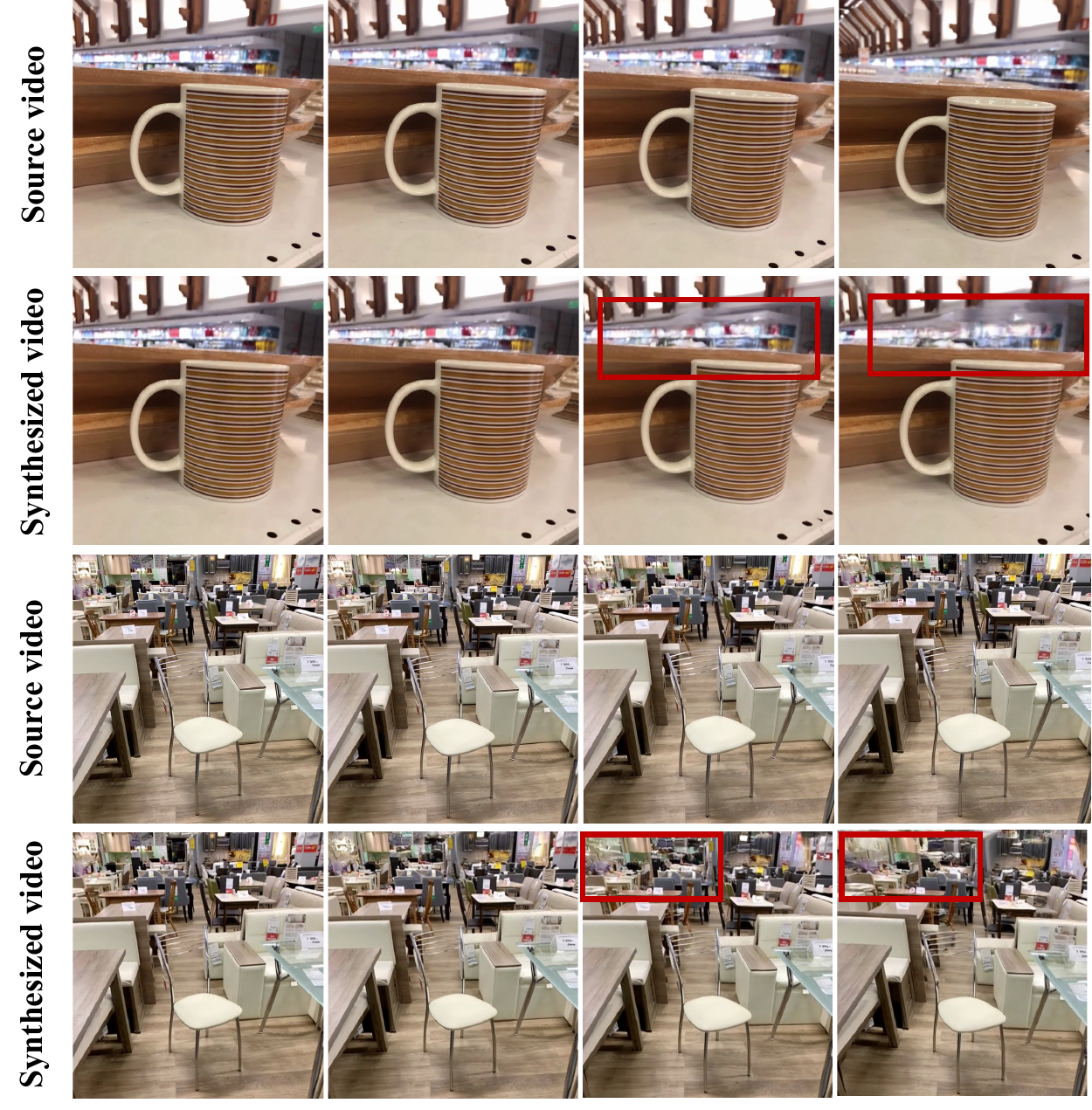}
  \caption{\textbf{Visualization of failure cases.}}
  \label{fig:suppl_failed_cases}
\end{figure}

We present the failure cases in Figure~\ref{fig:suppl_failed_cases}. Since our model is built upon a small-scale base model, it inherits certain limitations, particularly in complex scene synthesis. 
This is notably observed in the public subset, which features more intricate backgrounds. 
For instance, the model is prone to hallucinations when generating distant or ambiguous details, such as counters and complex furniture arrangements, as highlighted in the red boxes.

\section{Limitations} 
While ConsID-Gen achieves strong appearance preservation, several limitations merit further investigation. First, our method relies on a relatively small baseline due to resource constraints. Although it delivers clear improvements at this scale, adopting larger-capacity models (e.g., 14B) shows promising potential and constitutes an important direction for future work. Furthermore, the baseline is currently restricted to 81-frame sequences. While ConsID-Gen maintains high fidelity within this range, sustaining fine-grained visual consistency across substantially longer horizons remains an open challenge that we aim to address in future research.

\newpage

\begin{figure*}[!t]
\centering
\begin{tcolorbox}[
    colback=gray!3!white,
    colframe=gray!60!black,
    colbacktitle=gray!15!white,
    coltitle=black,
    fonttitle=\bfseries,
    boxrule=0.6pt,
    arc=2mm,
    left=2mm,
    right=2mm,
    top=1mm,
    bottom=1mm,
    width=\textwidth,
    sharp corners,
    title=\textbf{Instruction Template for Appearance-aware Captioning},
]
\textbf{System Prompt:}  
You are an expert video captioner for object-centric product and item videos.  
Your task is to write a short descriptive caption focusing only on the visible appearance of the main object.  

\begin{quote}
\textbf{Rules:}
\begin{itemize}[leftmargin=1.5em, topsep=2pt, itemsep=2pt]
    \item Write exactly 1–2 sentences under 40 words.
    \item Describe the main object with 5–7 attributes: category name, color or pattern, material and finish/texture (matte, glossy, brushed, woven), shape or form factor, size or scale cues (in hand, on table), notable parts/features (buttons, seams, zippers, ports, tread), any visible wear or defects, and transcribe any legible text or logos exactly as seen.
    \item Be strictly factual and concise. Use simple present tense.
    \item Do not include camera behavior, background context, or usage speculation.
    \item Focus only on the primary product. Mention hands or props only if they clarify size or material.
    \item Output only the caption text. No labels, prefixes, hashtags, or emojis.
\end{itemize}
\end{quote}

\medskip
\textbf{User Prompt:} Write 1–2 factual sentences describing the main item in the video. 
Include 5–7 visible attributes: category, color or pattern, material and finish/texture, shape or form factor, size or scale cues, notable features like seams or ports, any wear/defects, and exact readable text or logos. 
Keep it under 40 words, simple present tense, and output only the caption.

\end{tcolorbox}
\caption{\textbf{Instruction Template for Temporal-aware Captioning.}}
\label{box:template-appearance-caption}
\end{figure*}

\begin{figure*}[!t]
\centering
\begin{tcolorbox}[
    colback=gray!3!white,
    colframe=gray!60!black,
    colbacktitle=gray!15!white,
    coltitle=black,
    fonttitle=\bfseries,
    boxrule=0.6pt,
    arc=2mm,
    left=2mm,
    right=2mm,
    top=1mm,
    bottom=1mm,
    width=\textwidth,
    sharp corners,
    title=\textbf{Instruction Template for Appearance-aware Captioning},
]
\textbf{System Prompt:}  
You are an expert video captioner for object-centric videos.  
Your task is to generate a single coherent, factual, and detailed caption that integrates the provided appearance description of the main object into a natural temporal-aware observation.

\begin{quote}
\textbf{Rules:}
\begin{itemize}[leftmargin=1.5em, topsep=2pt, itemsep=2pt]
    \item Write 2–3 sentences under 60 words.
    \item Temporal dynamics must be the main focus: describe camera movement (type, direction, angle, framing), human interaction (holding, rotating, tapping, opening, placing), and object motion or state change (slides, flips, spins, opens, closes).
    \item Explicitly reuse at least 3–5 key details from the given appearance description (color, material, texture, shape, text, or distinctive features).
    \item Weave appearance details smoothly into the temporal description so the caption reads as one fluent observation, not separate parts.
    \item Use objective, factual language in simple present tense. Avoid subjective or aesthetic terms like “beautifully,” “showcases,” or “highlights.”
    \item Mention background or props only if directly relevant to scale or interaction.
    \item Output only the caption text. No labels, prefixes, hashtags, or emojis.
\end{itemize}
\end{quote}

\medskip
\textbf{User Prompt:}  
The main object’s appearance is: \ul{\{APPEARANCE\_DESCRIPTION\}}  
Write 1–2 factual and coherent sentences that integrate this appearance information naturally into a temporal-aware description.  
Explicitly reuse at least 3–5 details from the appearance description (color, material, texture, shape, text, distinctive features).  
Describe camera movement, angle, and framing, along with any human interaction or object motion, combining them in one fluent observation.  
Keep it concise, under 60 words, in simple present tense. Output only the caption.
\end{tcolorbox}
\caption{\textbf{Instruction Template for Temporal-aware Captioning.}}
\label{box:template-temporal-caption}
\end{figure*}

\end{document}